\documentclass{article}

    \usepackage[final]{neurips_2025}

\usepackage[utf8]{inputenc} 
\usepackage[T1]{fontenc}    
\usepackage{hyperref}       
\usepackage{url}            
\usepackage{booktabs}       
\usepackage{amsfonts}       
\usepackage{nicefrac}       
\usepackage{microtype}      
\usepackage{xcolor}         
\usepackage{natbib}
\setcitestyle{numbers,open={[},close={]}} 
\usepackage{graphicx}
\usepackage{multirow}
\usepackage{subfigure}
\usepackage{enumitem}
\usepackage{amsmath}
\usepackage{float}
\usepackage{colortbl}
\usepackage{algorithm}
\usepackage{algpseudocode}
\usepackage[compact]{titlesec}
\usepackage{makecell}
\usepackage[most]{tcolorbox}

\title{UniGTE: Unified Graph–Text Encoding for Zero-Shot Generalization across Graph Tasks and Domains}

\author{
  Duo Wang \quad Yuan Zuo\thanks{Corresponding author.} \quad Guangyue Lu \quad Junjie Wu \\
  MIIT Key Laboratory of Data Intelligence and Management, Beihang University \\
  \{wangduo58, zuoyuan, lugybuaa, wujj\}@buaa.edu.cn \\
}

\begin{document}

\maketitle

\begin{abstract}
Generalizing to unseen graph tasks without task-specific supervision is challenging: conventional graph neural networks are typically tied to a fixed label space, while large language models (LLMs) struggle to capture graph structure. We introduce \textbf{UniGTE}, an instruction-tuned encoder–decoder framework that unifies structural and semantic reasoning. The encoder augments a pretrained autoregressive LLM with learnable \emph{alignment tokens} and a \emph{structure-aware graph–text attention} mechanism, enabling it to attend jointly to a tokenized graph and a natural-language task prompt while remaining permutation-invariant to node order. This yields compact, \emph{task-aware graph representations}. Conditioned solely on these representations, a frozen LLM decoder \emph{predicts and reconstructs}: it outputs the task answer and simultaneously paraphrases the input graph in natural language. The reconstruction objective regularizes the encoder to preserve structural cues. UniGTE is instruction-tuned on five datasets spanning node-, edge-, and graph-level tasks across diverse domains, yet requires no fine-tuning at inference. It achieves new state-of-the-art zero-shot results on node classification, link prediction, graph classification and graph regression under cross-task and cross-domain settings, demonstrating that tight integration of graph structure with LLM semantics enables robust, transferable graph reasoning.
\end{abstract}

\section{Introduction}

Zero-shot learning in graph machine learning seeks to generalize to unseen tasks and domains without task-specific supervision. Although graph neural networks (GNNs) excel in fully supervised settings, they transfer poorly to new label spaces or data distributions without costly fine-tuning~\cite{ParetoGNN}. Inspired by recent progress in natural language processing (NLP), prompt-based extensions have been proposed to enhance GNN generalization~\cite{graphprompt,allinone}. However, the rigid architecture of conventional GNNs—especially their task-specific output heads—still hampers adaptability in zero-shot scenarios.

The advent of large language models (LLMs) opens new avenues for zero-shot reasoning on graphs. A direct approach serializes graph data into text and feeds it to an LLM~\cite{graph2text,gpt4graph,nlgraph,llm2graph}. While simple, this often underperforms because LLMs lack structural inductive bias~\cite{canllm}. Recent work therefore explores combining GNNs and LLMs, which can be grouped as follows.

\textbf{LLMs as enhancers.}  
These methods keep a GNN as the primary predictor and employ the LLM only to inject auxiliary semantic signals—for example, generating synthetic labels or textual node descriptions~\cite{natureisneed,llmgnn1,llmgnn2,labelfree}. Although such signals improve performance, the approaches inherit the architectural rigidity of GNNs and typically require retraining for new tasks. Replacing task-specific output heads with textual label embeddings enables limited zero-shot classification~\cite{OFA,zerog} but does not naturally extend to regression or other objectives, and semantic mismatch between graph and text remains an issue.

\textbf{LLMs as predictors.}  
Here the predictive role is assigned to the LLM, while a GNN supplies structural information aligned to the LLM’s semantic space—usually via self-supervised pretraining and cross-modal projection~\cite{graphgpt,unigraph,teaglm}. Because the two components are trained separately, it is hard to inject task-specific signals in a task-aware manner, limiting generalization. Deeper integrations such as GOFA~\cite{gofa} inject GNN features into LLM tokens at inference time, boosting zero-shot performance but at a high computational cost and with persistent cross-task and cross-domain gaps.

\medskip
\noindent\textbf{Our proposal.}  
We introduce \textbf{UniGTE}, a unified encoder–decoder framework that is instruction-tuned on a diverse suite of graph datasets (node classification, link prediction, graph classification; domains include citation networks, e-commerce graphs, molecular structures). The \emph{encoder} augments a pretrained autoregressive LLM to jointly consume a tokenized graph, a natural-language task prompt, and a fixed set of learnable \emph{alignment tokens}. During self-attention, these alignment tokens aggregate structural and prompt signals into a compact \emph{task-aware graph representation}. The \emph{decoder} is a frozen autoregressive LLM that conditions on this representation to (i) generate the task prediction and (ii) reconstruct the graph prompt, with the latter providing auxiliary supervision via a prompt-level loss. This design yields a single model that is permutation-invariant to node order, conditioned on task instructions, and capable of zero-shot generalization across modalities, tasks, and domains. Our key contributions are as follows:
\begin{itemize}[leftmargin=8pt]
    \item We present UniGTE—the first unified encoder–decoder architecture that achieves zero-shot generalization across diverse graph tasks and domains without any task-specific fine-tuning.
    \item UniGTE conditions graph representation learning on task prompts and embeds both graph structure and textual semantics in a common space, enabling flexible adaptation across modalities and objectives.
    \item Extensive experiments demonstrate state-of-the-art zero-shot results on node classification, link prediction, and graph regression across multiple domains.
\end{itemize}

\begin{figure}[t!]
    \centering
    \includegraphics[scale=0.42]{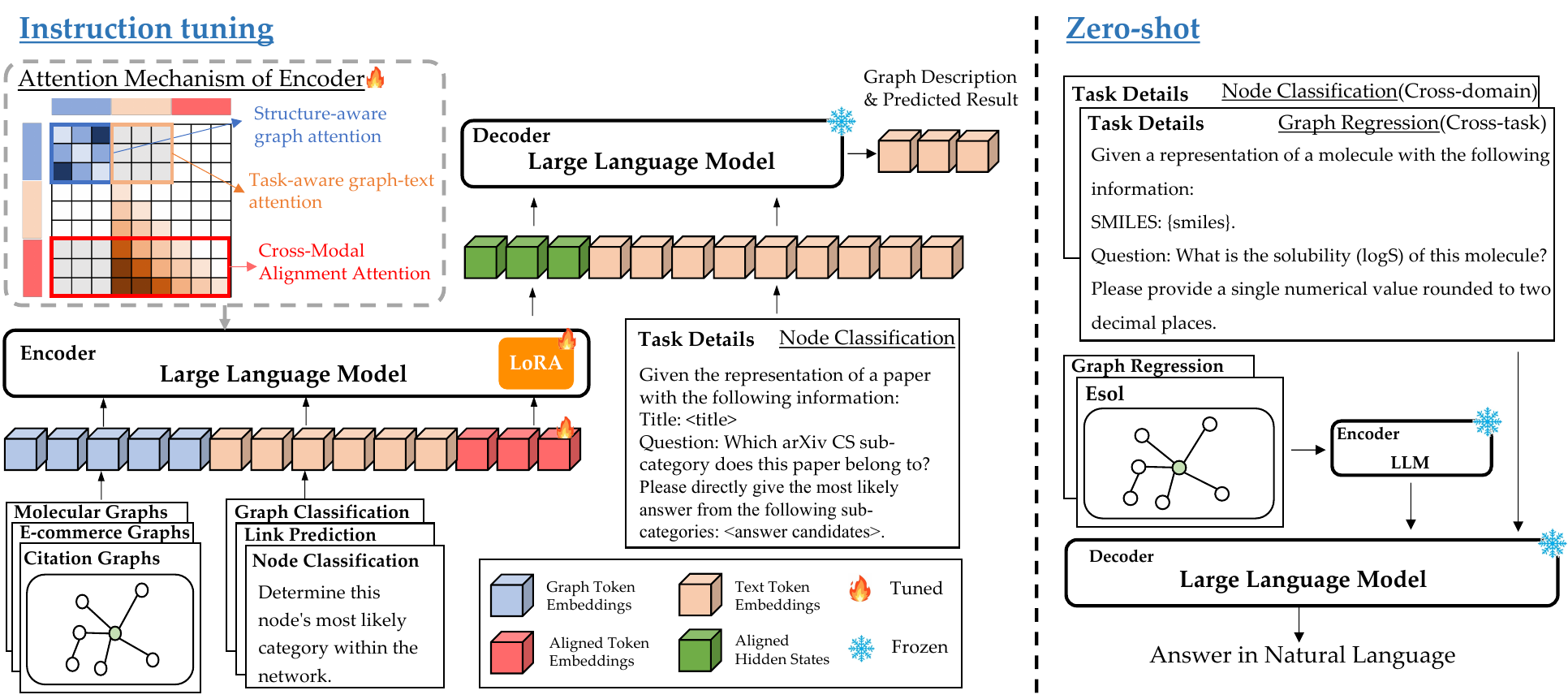}
        \caption{Framework of UniGTE}
    \label{fig:framework}
\end{figure}

\section{Methodology}
\label{sect:method}

We present \textbf{UniGTE}, a unified encoder–decoder framework for learning transferable graph representations across heterogeneous tasks and domains. UniGTE is instruction-tuned on a diverse collection of graph datasets that cover multiple task families—node classification, link prediction, and graph classification—and domains such as citation networks, e-commerce graphs, and molecular structures.

The architecture comprises an encoder and a decoder. The \emph{encoder} extends a pretrained autoregressive large language model (LLM) to jointly process graph-structured inputs and natural-language task prompts. Its input sequence comprises a tokenized graph (e.g., node representations from a subgraph), a task-specific prompt, and a fixed set of learnable \textit{alignment tokens}. These alignment tokens act as cross-modal anchors during self-attention, aggregating information from graph tokens guided by the task prompt and distilling it into a compact, task-aware latent representation—termed the \emph{task-aware graph representation}.

The \emph{decoder} is a frozen autoregressive LLM that conditions exclusively on the encoder outputs of the alignment tokens—that is, on the task-aware graph representation. It autoregressively generates two outputs: (i) the task prediction and (ii) a reconstruction of the graph prompt. The latter serves as an auxiliary supervision signal, implemented via a prompt-level loss that encourages the encoder to preserve the semantic content of the input graph. An overview of the framework is shown in Fig.~\ref{fig:framework}.

\subsection{Task definition and notation}

A graph-learning task $\tau \in \mathcal{T}$ is formally defined as $\tau := \bigl\{(\mathcal{G}_i, y_i^\tau)\bigr\}_{i=1}^{M} \cup \{T^\tau\}$, where $M$ is the number of graph instances. Each instance pairs a graph $\mathcal{G}_i$ with a target output $y_i^\tau$, expressed as a sequence of text tokens that typically encodes a class label or a numerical value.

A graph is denoted $\mathcal{G} = \bigl(\mathcal{V}, \mathcal{E}, \mathbf{A}, \mathbf{X}, \mathbf{E}\bigr)$, where $\mathcal{V} = \{v_1, v_2, \dots, v_{N}\}$ is the node set and $\mathcal{E} = \{e_1, e_2, \dots, e_{|\mathcal{E}|}\}$ the edge set. The adjacency matrix $\mathbf{A} \in \mathbb{R}^{N \times N}$ is defined by $\mathbf{A}_{ij}=1$ iff $(v_i, v_j)\in\mathcal{E}$. Node features are stored in $\mathbf{X} \in \mathbb{R}^{N \times F_N}$ and edge features in $\mathbf{E} \in \mathbb{R}^{|\mathcal{E}| \times F_E}$, with $F_N$ and $F_E$ denoting their respective feature dimensions.

Each task $\tau$ is accompanied by a natural-language instruction $T^\tau = [T^\tau_{\text{desc}},\,T^\tau_{\text{detail}}]$, containing two components. The task description $T^\tau_{\text{desc}} = [w_1^{(d)}, \dots, w_\ell^{(d)}]$ briefly states the objective—for example, \textit{“Determine this node's most likely category within the network's classification schema.”} The task-specific input $T^\tau_{\text{detail}} = [w_1^{(s)}, \dots, w_r^{(s)}]$ provides the contextualised information needed to solve the task, such as \textit{“Given a representation of a paper with the following information: Title: \{title\}, Abstract: \{abstract\}. Question: Which arXiv CS sub-category does this paper belong to?”}

\subsection{Unified graph-text encoder: learning task-aware graph representations via LLM}

In UniGTE, the encoder receives a graph instance $\mathcal{G}_i$, the task description $T^\tau_{\text{desc}}$, and a fixed set of alignment tokens $\mathcal{A}$ (defined below), and embeds them into a shared latent space. The decoder, conditioned on the encoder outputs corresponding to the alignment tokens—i.e., the task-aware graph representation—and on the task-specific input $T^\tau_{\text{detail}}$, then autoregressively produces the target sequence $y_i^\tau$.

We first describe the unified graph–text encoder. Prior work shows that graph tasks depend on structural and attribute information to different extents~\cite{nature,when}. To satisfy these varied requirements, the encoder must inject task cues into the aggregation process while retaining the ability to generalise across tasks and domains for zero-shot transfer. Inspired by the strong generalisation capacity of large language models (LLMs) in graph contexts, we build the encoder on a pretrained LLM that jointly encodes graph structure and task instructions, thereby learning task-aware representations.

\subsubsection{Unified input formatting for graph tasks}

To enable joint training across heterogeneous tasks and domains, we cast every task into a unified \emph{graph-level} input. For node- and edge-level tasks, we extract an $n$-hop subgraph centred on the target node or edge and treat all nodes in this subgraph as input tokens $T_{\mathcal{G}}$:
\[
T_{\mathcal{G}} = [w_1^{(g)}, \ldots, w_n^{(g)}],
\]
where each token $w_i^{(g)}$ is obtained by encoding the node’s attribute text with a pretrained language model (PLM), and $n = |\mathcal{G}|$ is the number of nodes in the subgraph. This abstraction lifts instance-level tasks to the graph level, promoting parameter sharing across task types~\cite{allinone}.

To inject task semantics, we employ a unified description template $T_{\text{desc}}$ that specifies both the task type and a concise summary of the input graph. Presenting this textual prompt alongside the graph tokens guides the encoder to produce a latent representation that is simultaneously structure- and task-aware.


To bridge graph and text inputs in a shared semantic space, we append a fixed set of learnable \emph{alignment tokens}:
\[
\mathcal{A} = [\,\mathbf{a}_1, \mathbf{a}_2, \ldots, \mathbf{a}_m\,], \qquad m \ll n + \ell,
\]
where $m$ is much smaller than the combined number of graph tokens $n$ and text tokens $\ell$. These tokens distil and integrate information from both modalities and act as the \emph{sole} interface to the decoder. This design unifies the representational spaces of graph and text while shrinking the overall token budget, allowing the model to accommodate longer inputs within its sequence-length limit.

Given $T_{\mathcal{G}}$, $T_{\text{desc}}$, and $\mathcal{A}$, the encoder input is
\[
\mathbf{x}_{\mathrm{enc}} =
    \bigl[\,
        T_{\mathcal{G}} \;\;;\;\;
        T_{\mathrm{desc}} \;\;;\;\;
        \mathcal{A}
    \,\bigr]
    \in \mathbb{R}^{(n + \ell + m)\times d_h},
\]
where $d_h$ is the hidden dimension of the underlying language model.

\subsubsection{Structure-aware graph-text attention}

Beyond simple input formatting, we propose a \emph{structure-aware graph--text attention} mechanism. Existing methods often linearize graph nodes into token sequences and input them into large language models (LLMs). However, their performance heavily relies on a \emph{fixed} node ordering and degrades significantly under permutation~\cite{wu2024can}. This fragility stems from the absolute positional encoding in standard self-attention, which contradicts the permutation invariance inherent to graph data. To address this, we design a unified attention mechanism that jointly processes graph and text tokens while preserving structural invariance.

Let the projection matrices be
\(
\mathbf{W}_Q, \mathbf{W}_K, \mathbf{W}_V \in \mathbb{R}^{d_h \times d_k}
\)
and
\(
\mathbf{W}_O \in \mathbb{R}^{d_k \times d_h}.
\)
The scaled dot-product attention with \emph{relative} position encoding, as applied in LLMs, is defined as
\begin{equation}
\hat{\mathbf{A}}_{ij}
=
\frac{(\mathbf{x}_i \mathbf{W}_Q)\,
       \mathbf{R}(i-j)\,
       (\mathbf{x}_j \mathbf{W}_K)^{\!\top}}
     {\sqrt{d_k}},\quad
\mathbf{A}=\operatorname{softmax}(\hat{\mathbf{A}}),\quad
\operatorname{Attn}(\mathbf{X})=\mathbf{A}(\mathbf{X}\mathbf{W}_V)\mathbf{W}_O,
\end{equation}
where \(\mathbf{R}(\cdot)\) denotes the rotary positional encoding (RoPE) transformation~\cite{rope}.

\paragraph{Cross-modal RoPE}

We extend RoPE to accommodate graph, text, and alignment tokens. For \emph{text} tokens, \(i\) and \(j\) represent absolute positions, and \(\mathbf{R}(i-j)\) operates as in standard RoPE. For \emph{graph} tokens, we assign a shared, learnable position index \texttt{<GraphPos>}, such that \(i-j=0\) for all graph-node pairs. This causes \(\mathbf{R}(0)\) to reduce to the identity matrix, removing order sensitivity. For cross-modal interactions, we compute the offset between text token positions and \texttt{<GraphPos>}, enabling the model to learn consistent alignments across modalities.

While this construction restores node-permutation invariance, it discards explicit structural cues. We re-inject them through additive biases:
\begin{equation}
\hat{\mathbf{A}}_{ij}
=
\frac{(\mathbf{x}_i \mathbf{W}_Q)\,
       \mathbf{R}(i-j)\,
       (\mathbf{x}_j \mathbf{W}_K)^{\!\top}}
     {\sqrt{d_k}}
\;+\;
b(i,j),
\end{equation}
with
\(
b(i,j)=
\mathbb{I}_{\{i,j\le n\}}\bigl(b^{\mathrm{PE}}_{ij}+b^{\mathrm{Edge}}_{ij}\bigr)
+b^{\mathrm{M}}_{ij}.
\)
The three bias terms are:

\paragraph{Distance-based structural bias}
Following graph transformers~\cite{gtf1}, we embed the shortest-path distance between graph nodes as:
\begin{equation}
b^{\mathrm{PE}}_{ij} = e\bigl(\operatorname{dist}_{\mathcal{G}}(i, j)\bigr),
\end{equation}
where \(e(\cdot)\) is a learnable lookup table, and \(\operatorname{dist}_{\mathcal{G}}(i, j)\) denotes the shortest-path distance between nodes \(i\) and \(j\).

\paragraph{Edge-aware bias}
Graphs in practice feature heterogeneous edge types. For each edge \(e_k\) on the shortest path \(\operatorname{SP}(i,j)\) we compute
\begin{equation}
b_{e_k}=\operatorname{MLP}\!\Bigl(\operatorname{PLM}\bigl(\text{``Description of }e_k\text{''}\bigr)\Bigr)
\end{equation}
and aggregate by
\(
b^{\mathrm{Edge}}_{ij}=\frac{1}{|\operatorname{SP}(i,j)|}\sum_{k\in\operatorname{SP}(i,j)}b_{e_k}.
\)
Unlike discrete type embeddings~\cite{gtf2}, this natural-language description lets the model generalize to unseen edge semantics, benefiting zero-shot transfer.

\paragraph{Masking bias}
Finally, \(b^{\mathrm{M}}_{ij}\) enforces directional constraints:
\begin{equation}
b^{\mathrm{M}}_{ij}=
\begin{cases}
-\infty,&
\text{if }(i>j\;\land\;i>n)
\;\text{or}\;
(i\le n\;\land\;n<j\le n+\ell),\\[2pt]
0,&\text{otherwise.}
\end{cases}
\end{equation}
Graph tokens attend bidirectionally to capture connectivity; in addition, they may look \emph{forward} into the text so downstream instructions can refine their representations. Conversely, text tokens never attend to graph tokens, preventing information leakage. Alignment tokens obey the standard causal mask across both modalities.

Together, these biases endow our attention with (i) permutation invariance for graph inputs, (ii) explicit structural awareness, and (iii) flexible cross-modal interaction, enabling robust reasoning across diverse graph-text tasks.


\subsubsection{Instruction tuning of the unified graph–text encoder}
\label{sec:it}

After the encoder processes the input graph and prompt, we take the hidden states $\mathcal{H_A}$ at the special \emph{alignment tokens}~$\mathcal{A}$ as task-aware graph representations.  
$\mathcal{H_A}$ fuse structural cues from the graph with semantic hints from the task description and constitute the sole conditioning signal for the decoder.

\paragraph{Instruction-tuning objective.}
We train concurrently on node-, edge- and graph-level tasks drawn from multiple domains.  
For a task~$\tau$ with target sequence ${\boldsymbol y}^\tau=(y_1^\tau,\dots,y_{L_\tau}^\tau)$ and instance-specific instruction $T^{\tau}_{\text{detail}}$, the negative log-likelihood loss is
\begin{equation}
\label{eq:it_loss}
\mathcal{L}_{\text{IT}}^\tau(\theta_{\text{enc}})
 \;=\;
 -\sum_{t=1}^{L_\tau}
 \log P\!\bigl(y_t^\tau \mid
               \mathcal{H_A},\,
               T^{\tau}_{\text{detail}},\,
               y_{<t}^\tau;\,
               \theta_{\text{enc}}\bigr).
\end{equation}
Here $T_{\text{detail}}^{\tau}$ differs from the terse task description $T_{\text{desc}}^{\tau}$ in that it contains the concrete, instance-level input needed to solve~$\tau$.

During instruction tuning we keep the decoder frozen and update only a small subset of encoder parameters: LoRA adapters $\theta_{\text{LoRA}}$, alignment-token embeddings $\theta_{\mathcal A}$, the MLP weights $\theta_{\text{MLP}}$ that compute edge-aware bias, and the table $\theta_{\text e}$ for relative-position bias.  We denote their union by~$\theta_{\text{enc}}$.

\paragraph{Auxiliary prompt reconstruction.}
We additionally ask the decoder to reconstruct the graph description $d_{\mathcal G}$ embedded in the prompt. This supervision encourages the alignment tokens to better encode structural information, while eliminating the need for a separate autoencoding stage. With target tokens $\boldsymbol w^{(d_{\mathcal G})}=(w^{(d_{\mathcal G})}_{1},\dots,w^{(d_{\mathcal G})}_{L_d})$, the auxiliary loss is
\begin{equation}
\label{eq:prompt_loss}
\mathcal{L}_{\text{prompt}}^\tau(\theta_{\text{enc}})
 \;=\;
 -\sum_{t=1}^{L_d}
 \log P\!\bigl(w_t^{(d_{\mathcal G})}\mid
               \mathcal{H_A},\,
               w_{<t}^{(d_{\mathcal G})};\,
               \theta_{\text{enc}}\bigr).
\end{equation}

\paragraph{Overall objective.}
The total loss for task~$\tau$ is the sum of the two terms:
\begin{equation}
\label{eq:total_loss}
\mathcal{L}_{\text{total}}^\tau(\theta_{\text{enc}})
 \;=\;
 \mathcal{L}_{\text{IT}}^\tau(\theta_{\text{enc}})
 \;+\;
 \mathcal{L}_{\text{prompt}}^\tau(\theta_{\text{enc}}).
\end{equation}
In training we minimise \(\sum_\tau \mathcal{L}_{\text{total}}^\tau(\theta_{\text{enc}})\) with respect to \(\theta_{\text{enc}}\).

\subsection{Training and evaluation strategy}
\label{train_strategy}

To assess the scalability and generalization ability of our model, we curate a diverse set of graph datasets spanning multiple levels (node, edge, and graph) and domains. The benchmark includes both classification and regression tasks drawn from application areas such as citation networks, e-commerce platforms, social media, and molecular graphs. Specifically, it comprises 17 datasets from five distinct domains, covering node classification, link prediction, graph classification, and graph regression tasks. Full details of the datasets are provided in the Appendix~\ref{sec:data_desc}. We use a subset of these datasets for instruction tuning, and directly evaluate the model's zero-shot performance on the remaining datasets without any further fine-tuning.

\section{Experimental results}
\label{sect:exper}
In this section, we conduct comprehensive experiments to validate the effectiveness of UniGTE. Our evaluation is designed to address the following research questions:

\begin{itemize}[leftmargin=28pt]
    \itemsep0em 
    \item[\textbf{RQ1:}] How well does UniGTE generalize to unseen datasets within the same domain (in-domain zero-shot)?
    \item[\textbf{RQ2:}] Can UniGTE handle more challenging generalization settings, such as transferring across domains or tasks unseen during training (cross-domain and cross-task zero-shot)?
    \item[\textbf{RQ3:}] What are the respective contributions of task-aware graph encoding and alignment tokens to the zero-shot performance of UniGTE?
\end{itemize}

\subsection{Experimental setup}
\label{sect:setup}

\paragraph{Datasets}
We jointly train UniGTE on five datasets: Arxiv~\cite{OGB}, Children~\cite{TAG}, Computer~\cite{TAG}, FB15K237~\cite{OFA}, and ChEMBL~\cite{chembl}, spanning node classification, link prediction, and graph classification tasks. For Arxiv, Children, and Computer, we construct both node classification and link prediction tasks to increase task diversity. After training, we evaluate UniGTE in a zero-shot setting on a set of unseen datasets. For in-domain evaluation, we use datasets from the same domains (e.g., additional citation or e-commerce graphs). To assess cross-domain generalization, we evaluate on datasets from different domains such as web graphs and social networks. Finally, to evaluate cross-task generalization, we include a previously unseen graph regression task. Detailed descriptions of all training and evaluation datasets are provided in the Appendix~\ref{sec:data_desc}.

\paragraph{Baselines}
We compare UniGTE against several recent state-of-the-art models with demonstrated transfer and zero-shot capabilities. \textbf{OFA}~\cite{OFA} combines a GNN-based predictor with a large language model (LLM) via prompt-based input augmentation. \textbf{GraphGPT}~\cite{graphgpt}, \textbf{LLaGA}~\cite{llaga}, and \textbf{TEA-GLM}~\cite{teaglm} adopt an LLM as the primary predictor and align graph-text representations through either a multi-layer perceptron (MLP) or a linear projection layer. \textbf{GOFA}~\cite{gofa} also employs an LLM as the predictor but incorporates structural information through inter-layer graph aggregation within the LLM architecture. Due to the high computational cost of training and evaluating \textbf{GraphGPT}, we report its results as provided in the original publication. For all other baselines, we re-ran the official implementations and conducted evaluations under our experimental setup. Detailed settings of the experimental environment can be found in Appendix~\ref{sec:exp_set}.

\subsection{In-domain zero-shot generalization (RQ1)}

To address RQ1, we evaluate each model’s zero-shot performance on datasets from the same domains as those seen during training. These include citation networks (\textbf{Pubmed}~\cite{Pubmed} and \textbf{Cora}~\cite{Cora}, with the latter being a more challenging variant featuring 70 classes), e-commerce datasets (\textbf{Photo} and \textbf{Sports}), and molecular graphs~\cite{mol}(\textbf{HIV}, \textbf{BACE}, and \textbf{PCBA}). We report accuracy for node classification, and AUC for link prediction and graph classification, reflecting the standard metrics used for each task type.

As shown in Table~\ref{tab:in_domain}, UniGTE achieves the best overall performance across tasks and datasets, outperforming all baselines on the majority of benchmarks.
Models relying on GNN-based predictors, such as OFA, struggle to generalize and exhibit weak transfer performance. Surprisingly, in most tasks, LLM-based methods like LLaGA and GraphGPT fail to outperform their base model, Vicuna-7B, suggesting that their lack of permutation invariance hinders generalization—changes in node ordering significantly impact their predictions.

Among the baselines, TEA-GLM applies a pooling mechanism to produce a fixed number of graph tokens, which preserves permutation invariance and contributes to better generalization in node classification and link prediction tasks. However, its inability to incorporate task-specific signals leads to inconsistent performance across tasks and even negative transfer in graph classification. GOFA, trained using our instruction tuning pipeline on the official pre-trained checkpoint, achieves limited gains. Despite extensive pretraining, it underperforms in most tasks and fails to match even an untuned LLM in many cases.

In contrast, UniGTE demonstrates consistent positive transfer across all datasets and task types. This can be attributed to its use of \textit{task-specific signals during graph encoding}, which enable the model to distinguish between task objectives, and its \textit{alignment tokens}, which unify structural and semantic information. Together, these components contribute to UniGTE’s superior generalization in in-domain, zero-shot scenarios.

\begin{table}[t!]
\centering
\caption{In-domain zero-shot results. \textbf{Bold} and \underline{underline} indicate the best and second-best results, respectively. N.S. denotes unsupported tasks.}
\label{tab:in_domain}
\resizebox{0.9\textwidth}{!}{
\begin{tabular}{c|cccc|ccc|cc}
\toprule
\multirow{2}{*}{\textbf{Model}} 
& \textbf{Pubmed} & \textbf{Cora} & \textbf{Photo} & \textbf{Sports} 
& \textbf{BACE} & \textbf{HIV} & \textbf{PCBA} 
& \textbf{Pubmed} & \textbf{Photo} \\
\cmidrule{2-10}
& \multicolumn{4}{c|}{Node Classification} & \multicolumn{3}{c|}{Graph Classification} & \multicolumn{2}{c}{Link Prediction} \\
\midrule
Vicuna-7B    & 0.721 & 0.155 & 0.384 & \underline{0.371} & 0.492 & 0.467 & 0.497 & 0.502 & 0.576 \\
OFA          & 0.237 & 0.189 & 0.317 & 0.047 & 0.483 & 0.404 & 0.424 & 0.499 & 0.499 \\
GraphGPT-std & 0.701 & 0.126 & --    & --    & --    & --    & --    & 0.501 & --    \\
LLaGA        & 0.726 & 0.156 & 0.249 & 0.351 & N.S.  & N.S.  & N.S.  & \textbf{0.740} & 0.659 \\
TEA-GLM      & \underline{0.781} & \underline{0.202} & 0.418 & 0.357 & 0.467 & \underline{0.498} & 0.434 & 0.663 & \underline{0.675} \\
GOFA         & 0.614 & 0.039 & \underline{0.447} & 0.133 & \underline{0.500} & 0.481 & \underline{0.500} & 0.507 & 0.504 \\
\midrule
\textbf{UniGTE} 
             & \textbf{0.870} & \textbf{0.215} & \textbf{0.565} & \textbf{0.403} 
             & \textbf{0.534} & \textbf{0.501} & \textbf{0.541} 
             & \underline{0.722} & \textbf{0.732} \\
\bottomrule
\end{tabular}
}
\end{table}

\subsection{Cross-domain and cross-task generalization (RQ2)}

To evaluate the generalization ability of each model in more challenging settings, we conduct zero-shot testing under both \textit{cross-domain} and \textit{cross-task} conditions. Specifically, we use datasets from domains not seen during training—\textbf{WikiCS}~\cite{wikics} (web links), \textbf{Reddit}, and \textbf{Instagram}~\cite{socialnet}(social networks)—as well as an entirely new task: \textit{graph regression}, evaluated on \textbf{Esol}~\cite{esol}, \textbf{Lipo}, and \textbf{Freesolv}~\cite{freesolv}. We report accuracy for node classification and mean absolute error (MAE) for regression.

As shown in Table~\ref{tab:cross}, UniGTE outperforms all baselines across both domains and task types. Most baseline models show some degree of positive transfer, but results remain inconsistent. LLaGA and OFA do not support graph-level tasks due to limitations in model design. GOFA achieves stronger regression performance than TEA-GLM on some datasets, yet both models show clear trade-offs: they perform well on specific domains or tasks but fail to generalize broadly.

In contrast, UniGTE consistently delivers strong performance across all datasets and settings. The challenging nature of cross-domain and cross-task transfer underscores the importance of robust generalization. UniGTE’s results demonstrate its ability to generalize effectively beyond the training distribution, highlighting the benefits of its unified graph-text representation and task-aware alignment mechanism.

\subsection{Ablation study (RQ3)}

We conduct ablation studies to assess the contributions of two key components in UniGTE: \textbf{alignment tokens} and \textbf{task-aware graph encoding}. To evaluate the role of alignment tokens, we remove them entirely and allow the decoder to generate outputs without their guidance. For task-aware graph encoding, we replace the task-specific description $T_{\text{desc}}$ with a fixed, generic prompt that is not tailored to any specific task.

To provide a comprehensive analysis, we evaluate performance from both \textbf{domain-level} and \textbf{task-level} perspectives, averaging metrics across datasets within each category. To ensure comparability across different task types—particularly between classification and regression—we adopt a normalized MAE score defined as:
\[
\widehat{\text{MAE}} = 1 - \frac{\text{MAE} - \text{MAE}_{\min}}{\text{MAE}_{\max} - \text{MAE}_{\min}},
\]
where $\text{MAE}_{\min}$ and $\text{MAE}_{\max}$ denote the minimum and maximum MAE values observed across all models and datasets. This normalization yields scores between 0 and 1, where higher values indicate better performance, thus making them directly comparable to metrics such as accuracy and AUC.

\begin{figure}[t!]
    \centering
    \subfigure[Domain-level perspective]{
        \label{ablation_nc}
        \includegraphics[scale=0.16]{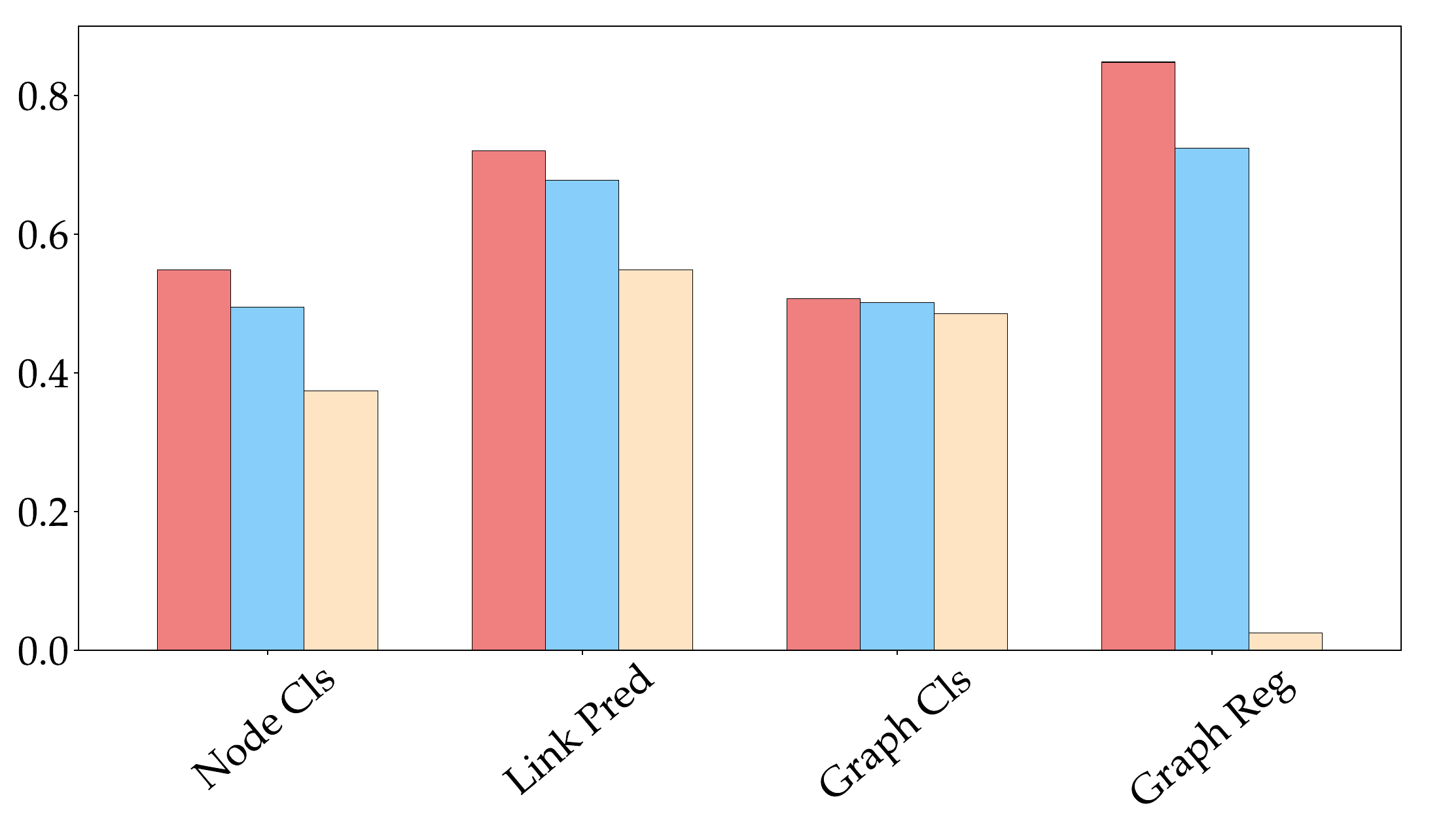}
    }
    \subfigure[Task-level perspective]{
        \label{ablation_lp}
        \includegraphics[scale=0.16]{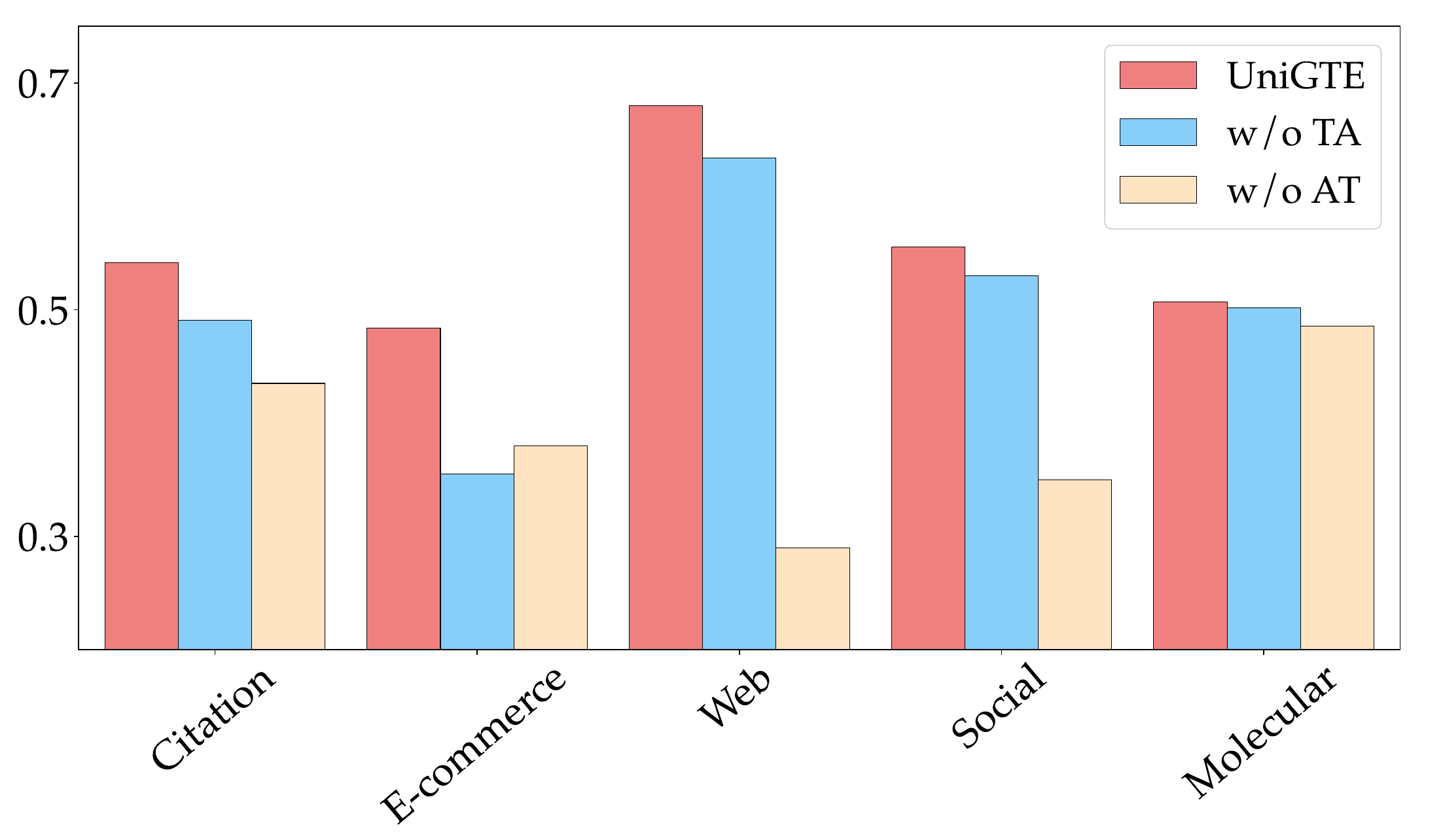}
    }
    \caption{
        Ablation study of alignment tokens and task-aware graph encoding, reported from both domain-level and task-level perspectives. ``w/o AT'' indicates the removal of alignment tokens, while ``w/o TA'' refers to replacing task-specific descriptions with a generic prompt.
    }
    \label{ablation}
\end{figure}

As shown in Figure~\ref{ablation}, both components contribute substantially to the model’s generalization. ``\textbf{w/o AT}'' corresponds to the setting without alignment tokens, and ``\textbf{w/o TA}'' reflects the ablation of task-aware graph encoding. Across both perspectives, removing either component results in a consistent performance drop. The absence of alignment tokens significantly degrades performance, highlighting their role in capturing structured and semantic information essential for zero-shot inference. Similarly, removing task-specific descriptions reduces performance across the board, confirming the importance of task-aware encoding in providing fine-grained task signals. 

Overall, these results underscore the necessity of both components: alignment tokens enhance generalization by bridging modalities, while task-aware encoding improves adaptability to diverse task objectives.

\begin{table}[t!]
\centering
\caption{Zero-shot performance on cross-domain node classification and cross-task graph regression. \textbf{Bold} and \underline{underline} indicate the best and second-best results, respectively. N.S. indicates tasks not supported by the model. Lower is better for regression (MAE).}
\label{tab:cross}
\resizebox{0.7\textwidth}{!}{
\begin{tabular}{c|ccc|ccc}
\toprule
\multirow{2}{*}{\textbf{Model}} 
& \textbf{WikiCS} & \textbf{Reddit} & \textbf{Instagram}
& \textbf{Esol} & \textbf{Lipo} & \textbf{Freesolv} \\
\cmidrule{2-7}
& \multicolumn{3}{c|}{Node Classification (Accuracy)} & \multicolumn{3}{c}{Graph Regression (MAE)} \\
\midrule
Vicuna-7B    & 0.290 & 0.309 & 0.391 & 6.58 & 11.22 & 64.11 \\
OFA          & 0.361 & 0.498 & \underline{0.580} & N.S. & N.S. & N.S. \\
LLaGA        & 0.601 & \underline{0.499} & 0.397 & N.S. & N.S. & N.S. \\
TEA-GLM      & 0.449 & 0.491 & 0.479 & 14.90 & 9.76 & \underline{13.35} \\
GOFA         & \underline{0.613} & 0.493 & 0.367 & \underline{4.93} & \underline{1.36} & 14.98 \\
\midrule
\textbf{UniGTE} 
             & \textbf{0.680} & \textbf{0.510} & \textbf{0.601}
             & \textbf{2.54} & \textbf{1.03} & \textbf{9.18} \\
\bottomrule
\end{tabular}}
\end{table}

\section{Related work}
\label{sec:related_work}

In this section, we review recent advances in zero-shot and transfer learning for graph machine learning. Existing approaches can be broadly categorized into two groups: (1) methods based solely on GNNs, and (2) methods that incorporate large language models (LLMs). The LLM-based methods can be further divided into two subtypes: \emph{LLM as an enhancer}, where the LLM provides auxiliary semantic information to assist graph models, and \emph{LLM as a predictor}, where the LLM itself performs prediction after being aligned with structural features.

\subsection{Zero-shot and transfer learning with GNNs}

A wide range of self-supervised learning techniques have been proposed to reduce the reliance of graph neural networks (GNNs) on labeled data and improve their generalization. For example, Deep Graph Infomax (DGI)~\cite{DGI} maximizes mutual information between local and global embeddings. Contrastive methods such as GraphCL~\cite{GraphCL}, GCA~\cite{GCA}, GCC~\cite{GCC}, and JOAO~\cite{JOAO} construct positive and negative views of graphs to learn invariant node representations. Generative approaches like GraphMAE~\cite{graphmae, graphmae2} adopt masked reconstruction objectives to jointly encode semantic and structural cues.

Although effective, these models typically require task-specific fine-tuning on downstream datasets. To improve transferability, prompt-based GNNs have recently gained attention. GraphPrompt~\cite{graphprompt} proposes a unified prompt template shared across pretraining and fine-tuning, improving knowledge transfer. ProG~\cite{allinone} reformulates node-level, link-level, and graph-level tasks into a unified prompting framework and leverages meta-learning for multi-task adaptation. However, these approaches still depend heavily on GNN-specific architectures, making it difficult to generalize across tasks or datasets with varying output spaces without further retraining.

\subsection{LLMs for graph zero-shot and transfer learning}
\subsubsection{LLMs as enhancers}
One line of work leverages LLMs as \emph{semantic enhancers}—generating task descriptions, pseudo-labels, or contextual cues to guide GNNs during training~\cite{llmgnn1, llmgnn2, labelfree, OFA,zerog}. These methods benefit from the rich prior knowledge encoded in LLMs; however, the final prediction is still performed by GNNs, limiting their flexibility and generalization in unseen tasks. Some studies~\cite{OFA,zerog} further propose novel architectural designs that enable GNNs to support zero-shot transfer across datasets. Despite these improvements, such approaches still exhibit limited generalization ability and are inherently unsuitable for regression tasks.

\subsubsection{LLMs as predictors}
A second line of work treats LLMs as \emph{predictors}. Some studies attempt to serialize graph data into textual sequences~\cite{gpt4graph,nlgraph,graph2text}, which are then directly fed into LLMs for inference. While this approach enables zero-shot reasoning by leveraging pretrained language models, it often compromises structural fidelity, as LLMs primarily capture token-level co-occurrence patterns rather than graph-specific inductive biases~\cite{canllm}.

Some methods like LLaGA~\cite{llaga} attempt to convert nodes into token sequences and input them directly into LLMs. However, this approach loses permutation invariance—a key property of graph data—making the output highly sensitive to node ordering~\cite{wu2024can}. To better preserve structural information, other works incorporate GNNs to extract graph features, which are then aligned with LLM inputs. For example, GraphGPT~\cite{graphgpt} aligns a graph transformer encoder with an LLM through a two-stage training process. UniGraph~\cite{unigraph} employs masked word prediction to pretrain a GNN and aligns it with LLM embeddings via an MLP-based projection. TEA-GLM~\cite{teaglm} further proposes a feature-wise contrastive pretraining strategy, followed by lightweight projection for alignment.vDespite these advances, most existing methods treat GNNs and LLMs as loosely coupled modules and perform alignment in a post hoc manner, limiting their adaptability to task-specific signals. As a result, such models often struggle in multi-task or cross-domain scenarios.

GOFA~\cite{gofa} addresses this limitation by introducing a tighter integration strategy, inserting GNN layers between LLM transformer blocks to enable token-level structural aggregation. However, this architecture incurs high computational costs and still faces performance challenges in zero-shot settings across diverse tasks and domains.

\section{Limitations}
\label{sect:limit}

While UniGTE achieves strong zero-shot generalization across a wide range of graph tasks and domains, its performance gains on link prediction tasks are less pronounced compared to node classification and graph regression. This may be due to the pairwise nature of link prediction, which poses unique challenges for prompt formulation and representation alignment. Exploring more effective strategies for encoding edge-level interactions and adapting task prompts for link prediction remains an important direction for future work.

\section{Conclusion}
\label{sect:conclusion}

We presented \textbf{UniGTE}, a unified encoder–decoder framework for zero-shot graph learning. UniGTE fuses graph structure and natural-language task instructions in a shared representation space via a permutation-invariant encoder with learnable alignment tokens, while a frozen LLM decoder produces both task predictions and prompt reconstructions. This design supports flexible transfer across node-, edge-, and graph-level tasks and across disparate domains. Extensive experiments verify UniGTE’s robustness, setting new zero-shot state-of-the-art results under demanding cross-task and cross-domain conditions. Our study highlights the benefit of tightly integrating structural and semantic cues for broadly transferable graph reasoning.

\section{Acknowledgement}
\label{sect:acknowledgement}

This work was supported by the National Key R\&D Program of China (2023YFC3304700). The work of Yuan Zuo was partially supported by the National Natural Science Foundation of China (72571019, 72531002) and the Shenzhen Science and Technology Program (CJGJZD20230724093201004). Dr. Junjie Wu’s work was partially supported by the National Natural Science Foundation of China (72031001, 72242101) and Outstanding Young Scientist Program of Beijing Universities (JWZQ20240201002).

\bibliography{neurips_2025}

\begin{thebibliography}{43}
\providecommand{\natexlab}[1]{#1}
\providecommand{\url}[1]{\texttt{#1}}
\expandafter\ifx\csname urlstyle\endcsname\relax
  \providecommand{\doi}[1]{doi: #1}\else
  \providecommand{\doi}{doi: \begingroup \urlstyle{rm}\Url}\fi

\bibitem[Casasnovas et~al.(2014)Casasnovas, Ortega-Castro, Frau, Donoso, and Munoz]{freesolv}
Rodrigo Casasnovas, Joaquin Ortega-Castro, Juan Frau, Josefa Donoso, and Francisco Munoz.
\newblock Theoretical pka calculations with continuum model solvents, alternative protocols to thermodynamic cycles.
\newblock \emph{International Journal of Quantum Chemistry}, 2014.

\bibitem[Chen et~al.(2024{\natexlab{a}})Chen, Zhao, Jaiswal, Shah, and Wang]{llaga}
Runjin Chen, Tong Zhao, Ajay Jaiswal, Neil Shah, and Zhangyang Wang.
\newblock Llaga: Large language and graph assistant.
\newblock In \emph{ICML}, 2024{\natexlab{a}}.

\bibitem[Chen et~al.(2023)Chen, Mao, Li, Jin, Wen, Wei, Wang, Yin, Fan, Liu, and Tang]{graph2text}
Zhikai Chen, Haitao Mao, Hang Li, Wei Jin, Hongzhi Wen, Xiaochi Wei, Shuaiqiang Wang, Dawei Yin, Wenqi Fan, Hui Liu, and Jiliang Tang.
\newblock Exploring the potential of large language models ({LLM}s) in learning on graph.
\newblock In \emph{NeurIPS 2023 Workshop: New Frontiers in Graph Learning}, 2023.
\newblock URL \url{https://openreview.net/forum?id=ScNNo7v4t0}.

\bibitem[Chen et~al.(2024{\natexlab{b}})Chen, Mao, Wen, Han, Jin, Zhang, Liu, and Tang]{labelfree}
Zhikai Chen, Haitao Mao, Hongzhi Wen, Haoyu Han, Wei Jin, Haiyang Zhang, Hui Liu, and Jiliang Tang.
\newblock Label-free node classification on graphs with large language models ({LLM}s).
\newblock In \emph{The Twelfth International Conference on Learning Representations}, 2024{\natexlab{b}}.
\newblock URL \url{https://openreview.net/forum?id=hESD2NJFg8}.

\bibitem[Gaulton et~al.(2011)Gaulton, Bellis, Bento, Chambers, Davies, Hersey, Light, McGlinchey, Michalovich, Al-Lazikani, and Overington]{chembl}
Anna Gaulton, Louisa~J. Bellis, A.~Patricia Bento, Jon Chambers, Mark Davies, Anne Hersey, Yvonne Light, Shaun McGlinchey, David Michalovich, Bissan Al-Lazikani, and John~P. Overington.
\newblock Chembl: a large-scale bioactivity database for drug discovery.
\newblock \emph{Nucleic Acids Research}, 2011.

\bibitem[Gomes et~al.(2022)Gomes, Ruelens, Efthymiadis, Nowe, and Vrancx]{when}
Diana Gomes, Frederik Ruelens, Kyriakos Efthymiadis, Ann Nowe, and Peter Vrancx.
\newblock When are graph neural networks better than structure-agnostic methods?
\newblock In \emph{I Can't Believe It's Not Better Workshop: Understanding Deep Learning Through Empirical Falsification}, 2022.

\bibitem[Guo et~al.(2023)Guo, Du, and Liu]{gpt4graph}
Jiayan Guo, Lun Du, and Hengyu Liu.
\newblock Gpt4graph: Can large language models understand graph structured data ? an empirical evaluation and benchmarking.
\newblock \emph{ArXiv}, abs/2305.15066, 2023.
\newblock URL \url{https://api.semanticscholar.org/CorpusID:258865990}.

\bibitem[He et~al.(2024)He, Bresson, Laurent, Perold, LeCun, and Hooi]{Pubmed}
Xiaoxin He, Xavier Bresson, Thomas Laurent, Adam Perold, Yann LeCun, and Bryan Hooi.
\newblock Harnessing explanations: {LLM}-to-{LM} interpreter for enhanced text-attributed graph representation learning.
\newblock In \emph{The Twelfth International Conference on Learning Representations}, 2024.
\newblock URL \url{https://openreview.net/forum?id=RXFVcynVe1}.

\bibitem[He and Hooi(2024)]{unigraph}
Yufei He and Bryan Hooi.
\newblock Unigraph: Learning a cross-domain graph foundation model from natural language.
\newblock \emph{arXiv preprint arXiv:2402.13630}, 2024.

\bibitem[Hou et~al.(2022)Hou, Liu, Cen, Dong, Yang, Wang, and Tang]{graphmae}
Zhenyu Hou, Xiao Liu, Yukuo Cen, Yuxiao Dong, Hongxia Yang, Chunjie Wang, and Jie Tang.
\newblock Graphmae: Self-supervised masked graph autoencoders.
\newblock In \emph{Proceedings of the 28th ACM SIGKDD Conference on Knowledge Discovery and Data Mining}, page 594–604, 2022.
\newblock URL \url{https://doi.org/10.1145/3534678.3539321}.

\bibitem[Hou et~al.(2023)Hou, He, Cen, Liu, Dong, Kharlamov, and Tang]{graphmae2}
Zhenyu Hou, Yufei He, Yukuo Cen, Xiao Liu, Yuxiao Dong, Evgeny Kharlamov, and Jie Tang.
\newblock Graphmae2: A decoding-enhanced masked self-supervised graph learner.
\newblock In \emph{Proceedings of the ACM Web Conference 2023}, page 737–746, 2023.
\newblock URL \url{https://doi.org/10.1145/3543507.3583379}.

\bibitem[Hu et~al.(2020)Hu, Fey, Zitnik, Dong, Ren, Liu, Catasta, and Leskovec]{OGB}
Weihua Hu, Matthias Fey, Marinka Zitnik, Yuxiao Dong, Hongyu Ren, Bowen Liu, Michele Catasta, and Jure Leskovec.
\newblock Open graph benchmark: Datasets for machine learning on graphs.
\newblock In \emph{Advances in Neural Information Processing Systems}, pages 22118--22133, 2020.
\newblock URL \url{https://proceedings.neurips.cc/paper_files/paper/2020/file/fb60d411a5c5b72b2e7d3527cfc84fd0-Paper.pdf}.

\bibitem[Huang et~al.(2023)Huang, Zhang, Mei, and Ma]{canllm}
Jin Huang, Xingjian Zhang, Qiaozhu Mei, and Jiaqi Ma.
\newblock Can llms effectively leverage graph structural information: when and why.
\newblock \emph{arXiv preprint arXiv:2309.16595}, 2023.

\bibitem[Ju et~al.(2023)Ju, Zhao, Wen, Yu, Shah, Ye, and Zhang]{ParetoGNN}
Mingxuan Ju, Tong Zhao, Qianlong Wen, Wenhao Yu, Neil Shah, Yanfang Ye, and Chuxu Zhang.
\newblock Multi-task self-supervised graph neural networks enable stronger task generalization.
\newblock In \emph{The Eleventh International Conference on Learning Representations}, 2023.
\newblock URL \url{https://openreview.net/forum?id=1tHAZRqftM}.

\bibitem[Kong et~al.(2025)Kong, Feng, Liu, Huang, Huang, Chen, and Zhang]{gofa}
Lecheng Kong, Jiarui Feng, Hao Liu, Chengsong Huang, Jiaxin Huang, Yixin Chen, and Muhan Zhang.
\newblock {GOFA}: A generative one-for-all model for joint graph language modeling.
\newblock In \emph{The Thirteenth International Conference on Learning Representations}, 2025.

\bibitem[Li et~al.(2024{\natexlab{a}})Li, Wang, Li, Yu, and Li]{zerog}
Yuhan Li, Peisong Wang, Zhixun Li, Jeffrey~Xu Yu, and Jia Li.
\newblock Zerog: Investigating cross-dataset zero-shot transferability in graphs.
\newblock \emph{arXiv preprint arXiv:2402.11235}, 2024{\natexlab{a}}.

\bibitem[Li et~al.(2024{\natexlab{b}})Li, Wang, Zhu, Chen, Jiang, Cai, Chan, and Li]{socialnet}
Yuhan Li, Peisong Wang, Xiao Zhu, Aochuan Chen, Haiyun Jiang, Deng Cai, Victor Wai~Kin Chan, and Jia Li.
\newblock Glbench: A comprehensive benchmark for graph with large language models, 2024{\natexlab{b}}.
\newblock URL \url{https://arxiv.org/abs/2407.07457}.

\bibitem[Liu and Wu(2023)]{llm2graph}
Chang Liu and Bo~Wu.
\newblock Evaluating large language models on graphs: Performance insights and comparative analysis.
\newblock \emph{arXiv preprint arXiv:2308.11224}, 2023.

\bibitem[Liu et~al.(2024)Liu, Feng, Kong, Liang, Tao, Chen, and Zhang]{OFA}
Hao Liu, Jiarui Feng, Lecheng Kong, Ningyue Liang, Dacheng Tao, Yixin Chen, and Muhan Zhang.
\newblock One for all: Towards training one graph model for all classification tasks.
\newblock In \emph{The Twelfth International Conference on Learning Representations}, 2024.
\newblock URL \url{https://openreview.net/forum?id=4IT2pgc9v6}.

\bibitem[Liu et~al.(2023)Liu, Yu, Fang, and Zhang]{graphprompt}
Zemin Liu, Xingtong Yu, Yuan Fang, and Xinming Zhang.
\newblock Graphprompt: Unifying pre-training and downstream tasks for graph neural networks.
\newblock In \emph{Proceedings of the ACM Web Conference 2023}, page 417–428, 2023.
\newblock URL \url{https://doi.org/10.1145/3543507.3583386}.

\bibitem[Mernyei and Cangea(2022)]{wikics}
Péter Mernyei and Cătălina Cangea.
\newblock Wiki-cs: A wikipedia-based benchmark for graph neural networks, 2022.
\newblock URL \url{https://arxiv.org/abs/2007.02901}.

\bibitem[Qiu et~al.(2020)Qiu, Chen, Dong, Zhang, Yang, Ding, Wang, and Tang]{GCC}
Jiezhong Qiu, Qibin Chen, Yuxiao Dong, Jing Zhang, Hongxia Yang, Ming Ding, Kuansan Wang, and Jie Tang.
\newblock Gcc: Graph contrastive coding for graph neural network pre-training.
\newblock In \emph{Proceedings of the 26th ACM SIGKDD International Conference on Knowledge Discovery \& Data Mining}, page 1150–1160, 2020.
\newblock URL \url{https://doi.org/10.1145/3394486.3403168}.

\bibitem[Su et~al.(2024)Su, Ahmed, Lu, Pan, Bo, and Liu]{rope}
Jianlin Su, Murtadha Ahmed, Yu~Lu, Shengfeng Pan, Wen Bo, and Yunfeng Liu.
\newblock Roformer: Enhanced transformer with rotary position embedding.
\newblock \emph{Neurocomputing}, 568:\penalty0 127063, 2024.

\bibitem[Sun et~al.(2023)Sun, Cheng, Li, Liu, and Guan]{allinone}
Xiangguo Sun, Hong Cheng, Jia Li, Bo~Liu, and Jihong Guan.
\newblock All in one: Multi-task prompting for graph neural networks.
\newblock In \emph{Proceedings of the 29th ACM SIGKDD Conference on Knowledge Discovery and Data Mining}, page 2120–2131, 2023.
\newblock URL \url{https://doi.org/10.1145/3580305.3599256}.

\bibitem[Tang et~al.(2023)Tang, Yang, Wei, Shi, Su, Cheng, Yin, and Huang]{graphgpt}
Jiabin Tang, Yuhao Yang, Wei Wei, Lei Shi, Lixin Su, Suqi Cheng, Dawei Yin, and Chao Huang.
\newblock Graphgpt: Graph instruction tuning for large language models.
\newblock \emph{arXiv preprint arXiv:2310.13023}, 2023.

\bibitem[Thang et~al.(2022)Thang, Dat, Tam, Jo, Hung, and Aberer]{nature}
Duong~Chi Thang, Hoang~Thanh Dat, Nguyen~Thanh Tam, Jun Jo, Nguyen Quoc~Viet Hung, and Karl Aberer.
\newblock Nature vs. nurture: Feature vs. structure for graph neural networks.
\newblock \emph{Pattern Recognition Letters}, 2022.

\bibitem[Veličković et~al.(2019)Veličković, Fedus, Hamilton, Liò, Bengio, and Hjelm]{DGI}
Petar Veličković, William Fedus, William~L. Hamilton, Pietro Liò, Yoshua Bengio, and R~Devon Hjelm.
\newblock Deep graph infomax.
\newblock In \emph{International Conference on Learning Representations}, 2019.
\newblock URL \url{https://openreview.net/forum?id=rklz9iAcKQ}.

\bibitem[Wang et~al.(2024)Wang, Zuo, Li, and Wu]{teaglm}
Duo Wang, Yuan Zuo, Fengzhi Li, and Junjie Wu.
\newblock Llms as zero-shot graph learners: Alignment of gnn representations with llm token embeddings.
\newblock In \emph{Advances in Neural Information Processing Systems}, 2024.

\bibitem[Wang et~al.(2023)Wang, Feng, He, Tan, Han, and Tsvetkov]{nlgraph}
Heng Wang, Shangbin Feng, Tianxing He, Zhaoxuan Tan, Xiaochuang Han, and Yulia Tsvetkov.
\newblock Can language models solve graph problems in natural language?
\newblock In \emph{Thirty-seventh Conference on Neural Information Processing Systems}, 2023.
\newblock URL \url{https://openreview.net/forum?id=UDqHhbqYJV}.

\bibitem[Wen and Fang(2023)]{Cora}
Zhihao Wen and Yuan Fang.
\newblock Augmenting low-resource text classification with graph-grounded pre-training and prompting.
\newblock In \emph{Proceedings of the 46th International ACM SIGIR Conference on Research and Development in Information Retrieval}, page 506–516, 2023.
\newblock \doi{10.1145/3539618.3591641}.
\newblock URL \url{https://doi.org/10.1145/3539618.3591641}.

\bibitem[Withnall et~al.(2018)Withnall, Chen, and Tetko]{esol}
Michael Withnall, Hongming Chen, and Igor~V Tetko.
\newblock Matched molecular pair analysis on large melting point datasets: a big data perspective.
\newblock \emph{ChemMedChem}, 2018.

\bibitem[Wu et~al.(2024)Wu, Shen, Shan, Song, Wang, Zhang, Feng, Cheng, Chen, Xiong, and Li]{wu2024can}
Xixi Wu, Yifei Shen, Caihua Shan, Kaitao Song, Siwei Wang, Bohang Zhang, Jiarui Feng, Hong Cheng, Wei Chen, Yun Xiong, and Dongsheng Li.
\newblock Can graph learning improve planning in {LLM}-based agents?
\newblock In \emph{The Thirty-eighth Annual Conference on Neural Information Processing Systems}, 2024.

\bibitem[Wu et~al.(2018)Wu, Ramsundar, Feinberg, Gomes, Geniesse, Pappu, Leswing, and Pande]{mol}
Zhenqin Wu, Bharath Ramsundar, Evan~N. Feinberg, Joseph Gomes, Caleb Geniesse, Aneesh~S. Pappu, Karl Leswing, and Vijay Pande.
\newblock Moleculenet: A benchmark for molecular machine learning, 2018.

\bibitem[Xia et~al.(2024)Xia, Kao, and Huang]{llmgnn2}
Lianghao Xia, Ben Kao, and Chao Huang.
\newblock Opengraph: Towards open graph foundation models.
\newblock \emph{arXiv preprint arXiv:2403.01121}, 2024.

\bibitem[Yan et~al.(2023)Yan, Li, Long, Yan, Zhao, Zhuang, Yin, Zhang, Han, Sun, Deng, Zhang, Sun, Xie, and Wang]{TAG}
Hao Yan, Chaozhuo Li, Ruosong Long, Chao Yan, Jianan Zhao, Wenwen Zhuang, Jun Yin, Peiyan Zhang, Weihao Han, Hao Sun, Weiwei Deng, Qi~Zhang, Lichao Sun, Xing Xie, and Senzhang Wang.
\newblock A comprehensive study on text-attributed graphs: Benchmarking and rethinking.
\newblock In \emph{Thirty-seventh Conference on Neural Information Processing Systems Datasets and Benchmarks Track}, 2023.
\newblock URL \url{https://openreview.net/forum?id=m2mbfoSuJ1}.

\bibitem[Ye et~al.(2023)Ye, Zhang, Wang, Xu, and Zhang]{natureisneed}
Ruosong Ye, Caiqi Zhang, Runhui Wang, Shuyuan Xu, and Yongfeng Zhang.
\newblock Natural language is all a graph needs.
\newblock \emph{arXiv preprint arXiv:2308.07134}, 2023.

\bibitem[Ying et~al.(2021{\natexlab{a}})Ying, Cai, Luo, Zheng, Ke, He, Shen, and Liu]{GraphCL}
Chengxuan Ying, Tianle Cai, Shengjie Luo, Shuxin Zheng, Guolin Ke, Di~He, Yanming Shen, and Tie-Yan Liu.
\newblock Do transformers really perform badly for graph representation?
\newblock In \emph{Advances in Neural Information Processing Systems}, pages 28877--28888, 2021{\natexlab{a}}.
\newblock URL \url{https://proceedings.neurips.cc/paper_files/paper/2021/file/f1c1592588411002af340cbaedd6fc33-Paper.pdf}.

\bibitem[Ying et~al.(2021{\natexlab{b}})Ying, Cai, Luo, Zheng, Ke, He, Shen, and Liu]{gtf2}
Chengxuan Ying, Tianle Cai, Shengjie Luo, Shuxin Zheng, Guolin Ke, Di~He, Yanming Shen, and Tie-Yan Liu.
\newblock Do transformers really perform badly for graph representation?
\newblock In \emph{Advances in Neural Information Processing Systems}, pages 28877--28888, 2021{\natexlab{b}}.
\newblock URL \url{https://proceedings.neurips.cc/paper_files/paper/2021/file/f1c1592588411002af340cbaedd6fc33-Paper.pdf}.

\bibitem[You et~al.(2021)You, Chen, Shen, and Wang]{JOAO}
Yuning You, Tianlong Chen, Yang Shen, and Zhangyang Wang.
\newblock Graph contrastive learning automated.
\newblock In \emph{ICML}, 2021.
\newblock URL \url{https://arxiv.org/abs/2106.07594}.

\bibitem[Yu et~al.(2023)Yu, Ren, Gong, Tan, Li, and Zhang]{llmgnn1}
Jianxiang Yu, Yuxiang Ren, Chenghua Gong, Jiaqi Tan, Xiang Li, and Xuecang Zhang.
\newblock Empower text-attributed graphs learning with large language models (llms).
\newblock \emph{arXiv preprint arXiv:2310.09872}, 2023.

\bibitem[Yun et~al.(2019)Yun, Jeong, Kim, Kang, and Kim]{gtf1}
Seongjun Yun, Minbyul Jeong, Raehyun Kim, Jaewoo Kang, and Hyunwoo~J Kim.
\newblock Graph transformer networks.
\newblock In \emph{Advances in Neural Information Processing Systems}, 2019.
\newblock URL \url{https://proceedings.neurips.cc/paper_files/paper/2019/file/9d63484abb477c97640154d40595a3bb-Paper.pdf}.

\bibitem[Zhang et~al.(2024)Zhang, Sun, Wang, Fan, Mo, Xu, Liu, Yang, and Shi]{graphtranslator}
Mengmei Zhang, Mingwei Sun, Peng Wang, Shen Fan, Yanhu Mo, Xiaoxiao Xu, Hong Liu, Cheng Yang, and Chuan Shi.
\newblock Graphtranslator: Aligning graph model to large language model for open-ended tasks.
\newblock In \emph{Proceedings of the ACM Web Conference 2023}, 2024.

\bibitem[Zhu et~al.(2021)Zhu, Xu, Yu, Liu, Wu, and Wang]{GCA}
Yanqiao Zhu, Yichen Xu, Feng Yu, Qiang Liu, Shu Wu, and Liang Wang.
\newblock Graph contrastive learning with adaptive augmentation.
\newblock In \emph{Proceedings of the Web Conference 2021}, page 2069–2080, 2021.
\newblock URL \url{https://doi.org/10.1145/3442381.3449802}.

\end{thebibliography}

\newpage
\appendix

\section{Dataset descriptions}
\label{sec:data_desc}

\begin{table}[H]
    \centering
    \caption{Dataset statistics}
    \label{dataset_table}
    \resizebox{0.8\textwidth}{!}{%
    \begin{tabular}{llrrrr}
    \toprule
    \textbf{Domain} & \textbf{Dataset} & \textbf{Avg. \#Nodes} & \textbf{Avg. \#Edges} & \textbf{\#Classes/Tasks} & \textbf{\#G}  \\
    \midrule
    \multirow[c]{3}{*}{Citation} & Arxiv & 169,343 & 1,166,243 & 40 & 1 \\
                              & Pubmed & 19,717 & 44,338 & 3 & 1 \\
                              & Cora & 25,120 & 91,140 & 70 & 1 \\
    \midrule
    \multirow[c]{4}{*}{E-commerce} & Computer & 87,229 & 721,081 & 10 & 1 \\
                                & Photo & 48,362 & 500,928 & 12 & 1 \\
                                & Children & 76,875 & 1,554,578 & 24 & 1 \\
                                & Sports & 173,055 & 1,773,500 & 13 & 1 \\
    \midrule
    Web link & WikiCS & 11,701 & 216,123 & 10 & 1 \\
    \midrule
    Knowledge graph & FB15K237 & 14,541 & 310,116 & 237 & 1 \\
    \midrule
    \multirow[c]{2}{*}{Social network} & Reddit & 33,434 & 198,438 & 2 & 1 \\
                                & Instagram & 11,339 & 144,010 & 2 & 1 \\
    \midrule
    \multirow[c]{7}{*}{Molecular} & ChEMBL & 25.87 & 55.92 & 1048 & 365,065 \\
                                & HIV & 25.51 & 54.95 & 1 & 41,127 \\
                                & BACE & 34.09 & 73.72 & 1 & 1,513 \\
                                & PCBA & 25.97 & 56.20 & 128 & 437,929 \\
                                & Esol & 13.29 & 27.35 & 1 & 1,128 \\
                                & Lipo & 27.04 & 59.00 & 1 & 4,200 \\
                                & Freesolv & 8.72 & 16.76 & 1 & 642 \\
    \bottomrule
    \end{tabular}}
\end{table}

\paragraph{Arxiv}
Arxiv~\cite{OGB} is a large-scale citation graph derived from arXiv Computer Science papers. Each node corresponds to a paper and edges represent citation links between papers. The task is to classify each paper into one of 40 arXiv subcategories, such as "cs.LG" or "cs.AI". This dataset serves as a representative benchmark for large-scale node classification.

\paragraph{Pubmed}
Pubmed~\cite{Pubmed} is a citation network of biomedical research papers from the PubMed database. Each node is a paper and edges correspond to citation links. The classification task involves assigning each paper to one of three disease-related categories.

\paragraph{Cora}
The Cora~\cite{Cora} dataset is a citation graph where each node corresponds to a research paper, and each edge represents a citation link between papers. The dataset focuses on papers within the machine learning domain and includes 70 fine-grained categories, making the classification task particularly difficult.

\paragraph{Computer}
The Computer~\cite{TAG} dataset is co-purchased or co-viewed product graph, where each node represents a product in the computer category, and edges indicate that two products were frequently co-purchased or co-viewed by users. The textual content associated with each node consists of user-generated reviews for the corresponding product.

\paragraph{Photo}
The Photo~\cite{TAG} dataset is an e-commerce product graph where nodes represent photographic products, and edges indicate that two items were either co-purchased or co-viewed by users. The textual content of each node consists of user reviews associated with the corresponding product.

\paragraph{Children}
The Children~\cite{TAG} dataset is a co-purchased or co-viewed product graph focused on children's books. Nodes correspond to individual books, and edges connect books that were frequently browsed or bought together. Each node is associated with textual information including the book's title and descriptive metadata.

\paragraph{Sports}
The Sports~\cite{TAG} dataset is a co-purchased or co-viewed product graph in the sports domain. Nodes represent sports-related products, and edges indicate that two items were often purchased or viewed together. The associated text for each node consists of the product's title.

\paragraph{WikiCS}
WikiCS~\cite{wikics} is a web link network constructed from English Wikipedia articles related to computer science. Nodes are individual articles, and directed edges represent hyperlinks between them. The node text is the full content of each article.

\begin{table}[t!]
\centering
\caption{Task-specific descriptions used as $T^\tau_{\text{desc}}$.}
\label{tab:task-desc}
\begin{tabular}{l p{7cm}}
\toprule
\textbf{Task Type} & \textbf{Task Description ($T^\tau_{\text{desc}}$)} \\
\midrule
Node Classification & Determine this node's most likely category \\
                         & within the network's classification schema. \\
\midrule
Link Prediction     & Determine whether there is a specific relationship \\
                         & between these two nodes. \\
\midrule
Graph Classification & Determine whether the molecule possesses specific \\
                          & physicochemical or bioactivity properties. \\
\midrule
Graph Regression     & Predict the continuous numerical value of a \\
                          & physicochemical or bioactivity property of the molecule. \\
\bottomrule
\end{tabular}
\end{table}

\paragraph{FB15K237}
FB15K237\cite{OFA} is a large-scale knowledge graph where each node represents an entity (e.g., a person, location, or object) and each edge corresponds to a relational triple connecting two entities. Textual content for nodes is constructed from entity names and relation descriptions.

\paragraph{Reddit}
The Reddit\cite{socialnet} dataset is a social interaction graph where nodes correspond to users and edges represent reply interactions in threads from a specific time period. Node-associated text consists of user posts from Reddit threads.

\paragraph{Instagram}
Instagram~\cite{socialnet} is a social graph in which each node represents a user, and edges denote social connections such as following relationships. The textual content associated with each node is extracted from users' self-introductions or profile descriptions.

\paragraph{ChEMBL}
ChEMBL~\cite{chembl} is a molecular graph dataset where each graph corresponds to a chemical compound. Nodes represent atoms, and edges denote chemical bonds. The textual information for each molecule is given by its SMILES (Simplified Molecular Input Line Entry System) representation.

\paragraph{HIV}
The HIV~\cite{mol} dataset consists of molecular graphs representing candidate compounds for HIV treatment. Nodes denote atoms and edges are chemical bonds. Each molecule is described by its SMILES string.

\paragraph{BACE}
BACE~\cite{mol} is a molecular dataset used in bioactivity classification. Each graph is a molecule, with atoms as nodes and chemical bonds as edges. SMILES strings provide the molecular structure information in text format.

\paragraph{PCBA}
PCBA~\cite{mol} is a large-scale molecular dataset for virtual screening. Each graph is a molecule, modeled by atoms and bonds, with SMILES strings representing the underlying chemical structure.

\paragraph{Esol}
The Esol~\cite{esol} dataset contains water-solubility data for chemical compounds. Each molecule is modeled as a graph, with node and edge structures corresponding to atoms and bonds. SMILES strings serve as the textual representation.

\paragraph{Lipo}
Lipo is a molecular dataset focused on lipophilicity prediction. Each molecule is represented as a graph with atoms as nodes and bonds as edges. The SMILES string encodes each molecule’s structure in text form.

\paragraph{Freesolv}
Freesolv~\cite{freesolv} consists of molecular graphs used for estimating hydration free energy. Each molecule is modeled by a graph of atoms and bonds. The SMILES representation is used as the text-based molecular description.

\begin{table}[t!]
\centering
\caption{Task-specific instruction templates used as $T^\tau_{\text{detail}}$.}
\label{tab:task-detail}
\renewcommand{\arraystretch}{1.2}
\resizebox{0.98\textwidth}{!}{
\begin{tabular}{l p{12.5cm}}
\toprule
\textbf{Dataset} & \textbf{Instruction Template ($T^\tau_{\text{detail}}$)} \\
\midrule
Arxiv, Pubmed, Cora & Given a representation of a paper with the following information: Title: \{title\}, Abstract: \{abstract\}. Question: Which arXiv CS sub-category does this paper belong to? Please directly give the most likely answer from the following sub-categories: \{candidate\_labels\}. \\
\midrule
Children & Given a representation of a book with the following information: Name: \{title\}, Content: \{abstract\}. Question: Which category does this book belong to? Please directly give the most likely answer from the following categories: \{candidate\_labels\}. \\
\midrule
Computer, Photo & Given a representation of a book with the following information: Name: \{title\}, Content: \{abstract\}. Question: Which category does this book belong to? Please directly give the most likely answer from the following categories: \{candidate\_labels\}. \\
\midrule
Sports & Given a representation of an electronic product with the following information: Comment: \{comment\}. Question: Which category does this electronic product belong to? Please directly give the most likely answer from the following categories: \{candidate\_labels\}. \\
\midrule
WikiCS & Given a representation of a Wikipedia page with the following information: Name: \{name\}, Content: \{content\}. Question: Which category does this Wikipedia page belong to? Please directly give the most likely answer from the following categories: \{candidate\_labels\}. \\
\midrule
Reddit & Given a representation of a user with the following information: Previous posts: \{posts\}. Question: Which category does this user belong to? Please directly give the most likely answer from the following categories: \{candidate\_labels\}. \\
\midrule
Instagram & Given a representation of a user with the following information: Personal introduction: \{introduction\}. Question: Which category does this user belong to? Please directly give the most likely answer from the following categories: \{candidate\_labels\}. \\
\midrule
FB15K237 & Given the representation of two entities: First entity: Name: \{name\}, Description: \{description\}. Second entity: Name: \{name\}, Description: \{description\}. Question: Which category should the relation between these two entities be classified as? Please directly give the most likely answer from the following categories: \{candidate\_labels\}. \\
\midrule
Arxiv-link, Pubmed-link & Given the representation of two papers: Title: First Paper: \{title\}, Second Paper: \{title\}. Question: Do these two papers have citation relationships? Please choose the most likely answer from: "Yes, they have citation relationships" or "No, they do not have citation relationships". \\
\midrule
Children-link & Given the representation of two books: Title: First Book: \{title\}, Second Book: \{title\}. Question: Do these two books have co-purchased or co-viewed relationships? Choose from: "Yes, they have co-purchased or co-viewed relationships" or "No, they do not have co-purchased or co-viewed relationships". \\
\midrule
Computer-link, Photo-link & Given the representation of two electronic products: Title: First Product: \{comment\}, Second Product: \{comment\}. Question: Do these two products have co-purchased or co-viewed relationships? Choose from: "Yes, they have co-purchased or co-viewed relationships" or "No, they do not have co-purchased or co-viewed relationships". \\
\midrule
ChEMBL, HIV, BACE, PCBA & Given a representation of a molecule with the following information: SMILES: \{smiles\}. Question: \{task\} Please answer: "Yes, this molecule is effective to this assay" or "No, this molecule is not effective to this assay". \\
\midrule
Esol, Lipo, Freesolv & Given a representation of a molecule with the following information: SMILES: \{smiles\}. Question: \{task\} Please provide a single numerical value rounded to two decimal places. \\
\bottomrule
\end{tabular}
}
\end{table}

\section{Details of prompt descriptions} 
\label{sec:prompt}

For each specific task $\tau$, we define a high-level textual description $T^\tau_{\text{desc}}$ that serves as the general instruction for the task. This description is task-type dependent and is designed to guide the model’s understanding and response generation. The full list of task descriptions used in our experiments is shown in Table~\ref{tab:task-desc}.

Additionally, we provide task-specific content instructions \(T^\tau_{\text{detail}}\), which include a description of the graph content and the corresponding question. The details are shown in the table~\ref{tab:task-detail}.

\section{Details of experimental setup}
\label{sec:exp_set}

\paragraph{Datasets} For data splitting, we follow standard splits for node classification, graph classification, and graph regression tasks. For link prediction, we randomly split the data into training/validation/test sets with a ratio of 8:1:1. To ensure fair comparison, all baseline models are evaluated using the same splits. Due to the large scale of the training data, we sample a subset of instances from each training dataset. Specifically, we sample 45,470 instances from \textbf{Arxiv}, 21,888 from \textbf{Children}, 31,378 from \textbf{Computer}, 10,000 each from \textbf{Arxiv-link}, \textbf{Children-link}, and \textbf{Computer-link}, 29,440 from \textbf{FB15K237}, and 74,242 from \textbf{ChEMBL}.

\paragraph{Baselines} For \textbf{LLaGA} and \textbf{TEA-GLM}, we re-run their training pipelines under our experimental settings using Vicuna-7B as the predictor. As \textbf{GOFA} requires extensive pretraining, we directly use the officially released pretrained checkpoint and conduct instruction tuning under our experimental setup. For all baselines, we follow the hyperparameter configurations provided in their respective original papers.

\paragraph{Training details} We use \textbf{BERT} as the pretrained language model (PLM), and \textbf{Vicuna-7B} as the large language model (LLM). In our implementation of UniGTE, we set the LoRA learning rate and the MLP learning rate to 2e-4, while the learning rate for the graph relative position embedding is 2e-3. 
We use a batch size of 2, apply gradient clipping with a maximum norm of 10, and perform gradient accumulation every 2 steps. 
The number of alignment tokens is fixed to 64 across all experiments. All experiments are conducted on a machine with two NVIDIA A100 GPUs, each equipped with 80GB of memory.

\section{More experimental results}
\subsection{Ablation study details}
\label{sec:ablation_detail}

\begin{table}[t!]
\centering
\caption{In-domain zero-shot results.}
\label{tab:ablation_in}
\resizebox{\textwidth}{!}{
\begin{tabular}{c|cccc|ccc|cc}
\toprule
\multirow{2}{*}{\textbf{Model}} 
& \textbf{Pubmed} & \textbf{Cora} & \textbf{Photo} & \textbf{Sports} 
& \textbf{BACE} & \textbf{HIV} & \textbf{PCBA} 
& \textbf{Pubmed} & \textbf{Photo} \\
\cmidrule{2-10}
& \multicolumn{4}{c|}{Node Classification} & \multicolumn{3}{c|}{Graph Classification} & \multicolumn{2}{c}{Link Prediction} \\
\midrule
w/o AT    & 0.721 & 0.155 & 0.384 & \underline{0.371} & \underline{0.492} & 0.467 & \underline{0.497} & 0.502 & 0.576 \\
w/o TA    & \underline{0.766} & \underline{0.211} & \underline{0.399} & 0.311 & 0.487 & \underline{0.499} & 0.489 & \underline{0.609} & \underline{0.710} \\
\midrule
\textbf{UniGTE} 
             & \textbf{0.870} & \textbf{0.215} & \textbf{0.565} & \textbf{0.403} 
             & \textbf{0.534} & \textbf{0.501} & \textbf{0.541} 
             & \textbf{0.722} & \textbf{0.732} \\
\bottomrule
\end{tabular}
}
\end{table}

\begin{table}[t!]
\centering
\caption{Zero-shot performance on cross-domain node classification and cross-task graph regression.}
\label{tab:ablation_cross}
\begin{tabular}{c|ccc|ccc}
\toprule
\multirow{2}{*}{\textbf{Setting}} 
& \textbf{WikiCS} & \textbf{Reddit} & \textbf{Instagram}
& \textbf{Esol} & \textbf{Lipo} & \textbf{Freesolv} \\
\cmidrule{2-7}
& \multicolumn{3}{c|}{Node Classification (Accuracy)} & \multicolumn{3}{c}{Graph Regression (MAE)} \\
\midrule
w/o AT    & 0.290 & 0.309 & 0.391 & \underline{6.58} & 11.22 & 64.11 \\
w/o TA          & \underline{0.645} & \underline{0.502} & \underline{0.590} & 8.35 & \underline{2.44} & \underline{12.37} \\
\midrule
\textbf{UniGTE} 
             & \textbf{0.680} & \textbf{0.510} & \textbf{0.601}
             & \textbf{2.54} & \textbf{1.03} & \textbf{9.18} \\
\bottomrule
\end{tabular}
\end{table}

We provide detailed ablation results for each dataset. The in-domain results are presented in Table~\ref{tab:ablation_in}, while the cross-domain and cross-task results are shown in Table~\ref{tab:ablation_cross}. As the tables indicate, removing the alignment tokens significantly degrades the model's transferability. Additionally, replacing the task-specific description with a fixed, generic prompt leads to performance drops across different datasets. These findings validate the effectiveness of the proposed alignment tokens in enhancing generalization, and further demonstrate the importance of incorporating task-specific signals into the encoding process.

\subsection{Legality rate}
\label{sec:legality}

\begin{table}[t!]
    \centering
    \caption{Legality rate (\%) across models and datasets}
    \resizebox{\textwidth}{!}{%
    \begin{tabular}{c|ccccccccccccc}
    \toprule
    \textbf{Dataset} & \textbf{Pubmed} & \textbf{Cora} & \textbf{Photo} & \textbf{Sports} & \textbf{WikiCS} & \textbf{Reddit} & \textbf{Instagram} & \textbf{HIV} & \textbf{BACE} & \textbf{PCBA} & \textbf{Esol} & \textbf{Lipo} & \textbf{Freesolv} \\
    \midrule
    Model & \multicolumn{13}{c}{Legality rate (\%)} \\
    \midrule
    Vicuna-7B & \textbf{100} & \textbf{95.8} & 94.1 & 99.6 & 64.7 & 57.4 & 90.3 & \underline{96.0} & 92.1 & 39.0 & \underline{99.1} & 81.2 & \underline{98.5} \\
    LLaGA     & \textbf{100} & 76.5 & 83.8 & 99.2 & \underline{99.0} & \underline{99.0} & \underline{99.5} & 82.1 & \textbf{100} & 75.1 & 66.3 & 82.2 & 57.0 \\
    TEA-GLM   & \textbf{100} & \underline{95.6} & \underline{96.2} & \textbf{100} & 70.9 & 96.8 & 99.7 & \textbf{100} & \textbf{100} & \textbf{100} & 80.6 & \underline{88.1} & 45.0 \\
    \midrule
    UniGTE    & \textbf{100} & 95.0 & \textbf{99.7} & \underline{99.7} & \textbf{99.6} & \textbf{100} & \textbf{100} & \textbf{100} & \textbf{100} & \textbf{100} & \textbf{99.1} & \textbf{99.8} & \textbf{100} \\
    \bottomrule
    \end{tabular}
    }
    \label{tab:legality_rate_models}
\end{table}

The training process may affect the instruction-following ability of LLMs in zero-shot scenarios. Specifically, while the model can generate appropriate outputs on the training set, it often fails to produce legality answers on unseen datasets. This essentially reflects poor generalization ability. Following the approach proposed in~\cite{graphtranslator}, we use the \textit{legality rate} to measure the proportion of valid responses generated by the model on unseen datasets.

As shown in Table~\ref{tab:legality_rate_models}, existing baselines exhibit poor instruction-following performance on certain datasets. In contrast, our model consistently maintains strong instruction compliance across all unseen datasets, generating valid answers. This further demonstrates the superior generalization capability of our approach.

\subsection{Efficiency and computational complexity analysis}
\label{sec:efficiency}

We have conducted a comparison between LLM and UniGTE under a zero-shot setting using equivalent graph information. The results are summarized in the tables below, where the first table reports inference time and the second table presents task-specific performance metrics, showing that UniGTE achieves significantly faster inference speed while maintaining superior accuracy.

\begin{table}[H]
    \centering
    \caption{Inference time compared with the LLM}
    \resizebox{\textwidth}{!}{%
    \begin{tabular}{c|ccccccc}
    \toprule
    \textbf{Sec / sample (s)} & \textbf{Pubmed} & \textbf{Cora} & \textbf{PCBA} & \textbf{Photo\_link} & \textbf{WikiCS} & \textbf{Instagram} & \textbf{Lipo} \\
    \midrule
    Vicuna-7B  & 0.8 & 2.6 & 3.2 & 1.7 & 0.7 & 3.3 & 10.2 \\
    UniGTE  & 0.5 & 0.6 & 1.0 & 1.1 & 0.3 & 0.3 & 2.0 \\
    \midrule
    Improvement & 37.5\% & 78.7\% & 70.3\% & 34.1\% & 50.8\% & 91.7\% & 80.5\% \\
    \bottomrule
    \end{tabular}
    \label{tab:llm_inference_time}
    }
\end{table}

\begin{table}[H]
    \centering
    \caption{Task performance compared with the LLM}
    \resizebox{\textwidth}{!}{%
    \begin{tabular}{c|ccccccc}
    \toprule
    \textbf{Task Metric} & \textbf{Pubmed} & \textbf{Cora} & \textbf{PCBA} & \textbf{Photo\_link} & \textbf{WikiCS} & \textbf{Instagram} & \textbf{Lipo} \\
    \midrule
    Vicuna-7B  & 0.721 & 0.155 & 0.497 & 0.576 & 0.290 & 0.391 & 11.220 \\
    UniGTE  & 0.870 & 0.215 & 0.541 & 0.732 & 0.680 & 0.601 & 1.030 \\
    \midrule
    Improvement & 20.7\% & 38.7\% & 8.9\% & 27.1\% & 134.5\% & 53.7\% & 90.8\% \\
    \bottomrule
    \end{tabular}
    \label{tab:llm_task_performance}
    }
\end{table}

Here, we compare the time complexity of GOFA. We adopt the same notation used in the GOFA paper: suppose a graph contains $V$ nodes, $E$ edges, and each node is represented by $k$ tokens on average. In GOFA, each node is expanded into $k$ tokens ($k \gg 1$), and the self-attention is restricted within each node's token group. Thus, the per-layer time complexity is: $\mathcal{O}(k \cdot k \cdot V) = \mathcal{O}(Vk^2)$, which is consistent with the complexity reported in the original GOFA paper. In contrast, UniGTE represents each node using a single token and performs cross-node attention directly. Therefore, the complexity becomes: $\mathcal{O}((V \cdot 1)^2) = \mathcal{O}(V^2)$. This is considerably lower, especially when $k = 128$ or higher as required in GOFA for satisfactory performance. Furthermore, our method leverages $k$-hop subgraph sampling, which keeps the number of nodes $V$ relatively small, making $\mathcal{O}(V) \ll \mathcal{O}(k^2)$ and thus $\mathcal{O}(V^2) \ll \mathcal{O}(Vk^2)$.

In addition to the Transformer layers of the LLM itself, GOFA also introduces external GNN layers. This component is not accounted for in its original complexity analysis. Assuming the average node degree is $d$, the GNN layer introduces an additional cost of $\mathcal{O}(V \cdot d \cdot k)$. Combining all these considerations, our method provides a clear computational advantage over GOFA.

To further validate this, we compare per-sample inference time and accuracy under the zero-shot setting between GOFA and UniGTE. As shown in the table below, UniGTE achieves faster inference and better performance, demonstrating the efficiency and effectiveness of our design.

\begin{table}[H]
    \centering
    \caption{Inference time compared with GOFA}
    \resizebox{\textwidth}{!}{%
    \begin{tabular}{c|ccccccc}
    \toprule
    \textbf{Sec / sample (s)} & \textbf{Pubmed} & \textbf{Cora} & \textbf{PCBA} & \textbf{Photo\_link} & \textbf{WikiCS} & \textbf{Instagram} & \textbf{Lipo} \\
    \midrule
    GOFA   & 1.4 & 5.1 & 1.5 & 1.8 & 2.0 & 1.2 & 2.1 \\
    UniGTE & 0.5 & 0.6 & 1.0 & 1.1 & 0.3 & 0.3 & 2.0 \\
    \midrule
    Improvement & 63.2\% & 89.3\% & 34.5\% & 38.5\% & 83.6\% & 77.1\% & 5.7\% \\
    \bottomrule
    \end{tabular}
    \label{tab:gofa_inference_efficiency}
    }
\end{table}

\begin{table}[t!]
    \centering
    \caption{Task performance compared with GOFA}
    \resizebox{\textwidth}{!}{%
    \begin{tabular}{c|ccccccc}
    \toprule
    \textbf{Task Metrics} & \textbf{Pubmed} & \textbf{Cora} & \textbf{PCBA} & \textbf{Photo\_link} & \textbf{WikiCS} & \textbf{Instagram} & \textbf{Lipo} \\
    \midrule
    GOFA   & 0.614 & 0.039 & 0.500 & 0.504 & 0.613 & 0.367 & 1.360 \\
    UniGTE & 0.870 & 0.215 & 0.541 & 0.732 & 0.680 & 0.601 & 1.030 \\
    \midrule
    Improvement & 41.7\% & 451.3\% & 8.2\% & 45.2\% & 10.9\% & 63.8\% & 24.3\% \\
    \bottomrule
    \end{tabular}
    \label{tab:gofa_task_performance}
    }
\end{table}

\subsection{Supervised results}
We have conducted additional experiments under the SFT setting and compared UniGTE with relevant baselines that use LLMs as predictors. The results are presented below.

As shown in Table~\ref{tab:sft}, UniGTE achieves competitive performance and yields strong results on most datasets, even surpassing state-of-the-art baselines on several of them. This highlights its promising capability under SFT settings. These results demonstrate that although our model is originally designed for zero-shot scenarios, it also performs robustly under full supervision, without significant degradation compared to existing approaches.

\begin{table}[t!]
    \centering
    \caption{Supervised results under SFT setting}
    \resizebox{\textwidth}{!}{%
    \begin{tabular}{c|ccccccc}
    \toprule
    \textbf{Model} & \textbf{Arxiv} & \textbf{Children} & \textbf{Computer} & \textbf{Arxiv\_link} & \textbf{Children\_link} & \textbf{Computer\_link} & \textbf{FB15K237} \\
    \midrule
    GraphGPT-std & 0.625 & - & - & - & - & - & - \\
    LLaGA & \underline{0.711} & \textbf{0.498} & \underline{0.579} & \underline{0.901} & \textbf{0.837} & \textbf{0.793} & 0.870 \\
    TEA-GLM & 0.661 & \underline{0.477} & 0.549 & 0.891 & 0.767 & 0.738 & \textbf{0.910} \\
    GOFA & 0.659 & 0.190 & 0.514 & 0.772 & 0.505 & 0.563 & 0.731 \\
    \midrule
    \textbf{UniGTE} & \textbf{0.713} & 0.431 & \textbf{0.589} & \textbf{0.913} & \underline{0.813} & \underline{0.765} & \underline{0.901} \\
    \bottomrule
    \end{tabular}
    \label{tab:sft}
    }
\end{table}

\subsection{Graph understanding tasks}

We have conducted additional experiments on graph understanding tasks to demonstrate our model’s graph understanding/reasoning capabilities, and compared the results with relevant baselines.

\begin{table}[H]
    \centering
    \caption{Comparison of various models on graph reasoning tasks}
    \resizebox{\textwidth}{!}{%
    \begin{tabular}{c|ccccccccc}
    \toprule
    \textbf{Model} & \textbf{Pubmed\_CONN} & \textbf{Cora\_CONN} & \textbf{Pubmed\_SPD} & \textbf{Cora\_SPD} & \textbf{Pubmed\_CN} & \textbf{Cora\_CN} & \textbf{BACE\_CYCLE} & \textbf{HIV\_CYCLE} & \textbf{PCBA\_CYCLE} \\
    \midrule
    Vicuna-7B & 0.551 & 0.505 & 1.51 & 2.53 & 5.07 & 12.85 & \underline{2.04} & \underline{1.37} & \underline{2.11} \\
    LLaGA & 0.618 & 0.553 & 3.72 & 7.98 & - & - & - & - & - \\
    TEA-GLM & \underline{0.689} & \underline{0.623} & 3.06 & 2.34 & 5.27 & 7.79 & 6.94 & 7.39 & 9.74 \\
    GOFA & 0.638 & \textbf{0.694} & \underline{1.43} & \underline{1.68} & \textbf{1.13} & \underline{7.24} & 6.96 & 5.58 & 6.24 \\
    \midrule
    \textbf{UniGTE} & \textbf{0.707} & 0.616 & \textbf{1.11} & \textbf{1.42} & \underline{1.20} & \textbf{2.24} & \textbf{1.83} & \textbf{1.10} & \textbf{1.92} \\
    \bottomrule
    \end{tabular}
    }
    \label{tab:grt}
\end{table}

We directly evaluated the model trained in our original submission on zero-shot graph understanding tasks, without any fine-tuning on these tasks. All other baselines were evaluated under the same zero-shot setting, except for GOFA, which incorporates some graph understanding tasks during its pretraining phase. To ensure fair comparison, all methods were tested on the same test sets. Following the experimental setup of GOFA and the references you kindly suggested, we constructed the following tasks and generated the data using real-world datasets. We used AUC for evaluating the CONN task, and MAE for the other tasks.

\begin{itemize}[leftmargin=8pt]
    \item CONN: Determine whether two nodes are connected
    \item SPD: Predict the shortest path distance between two nodes
    \item CN: Predict the number of common neighbors between two nodes
    \item CYCLE: Predict the number of cycles in the graph
\end{itemize}

The results are summarized in the Table~\ref{tab:grt}, where the best result is bolded and the second-best is italicized. Our method achieves state-of-the-art performance on most datasets. Notably, compared to Vicuna, which serves as our base model, our method consistently outperforms it across all tasks and datasets—highlighting the strong generalization and zero-shot capabilities of our approach. Compared to other baselines, our method is either the best or competitive. It is worth noting that GOFA, benefiting from the inclusion of graph understanding tasks during its pretraining phase, performs relatively well on a few specific datasets. In contrast, our method has never encountered any graph understanding tasks during training, making the evaluation a significantly more challenging and rigorous zero-shot setting. Despite this, our approach still delivers consistently strong performance, further demonstrating its generalization capability.


\newpage
\section*{NeurIPS Paper Checklist}

\begin{enumerate}

\item {\bf Claims}
    \item[] Question: Do the main claims made in the abstract and introduction accurately reflect the paper's contributions and scope?
    \item[] Answer: \answerYes{} 
    \item[] Justification: We verify the contributions of our proposed method through experiments, and the results in the Sec.~\ref{sect:exper} effectively demonstrate the contributions we outlined in the abstract and introduction.

\item {\bf Limitations}
    \item[] Question: Does the paper discuss the limitations of the work performed by the authors?
    \item[] Answer: \answerYes{} 
    \item[] Justification: We discussed the current shortcomings of our method and future research directions in Sec.\ref{sect:limit}

\item {\bf Theory assumptions and proofs}
    \item[] Question: For each theoretical result, does the paper provide the full set of assumptions and a complete (and correct) proof?
    \item[] Answer: \answerNA{} 
    \item[] Justification: The paper does not include theoretical results. 

    \item {\bf Experimental result reproducibility}
    \item[] Question: Does the paper fully disclose all the information needed to reproduce the main experimental results of the paper to the extent that it affects the main claims and/or conclusions of the paper (regardless of whether the code and data are provided or not)?
    \item[] Answer: \answerYes{} 
    \item[] Justification: We provide all necessary information to ensure the reproducibility of our main experimental results. Section~\ref{sect:method} details the proposed methodology, while Sections~\ref{train_strategy} and detailed setting in Appendix~\ref{sec:exp_set} outline the training procedures and implementation settings. All reported results in this paper can be reliably reproduced based on the disclosed configurations.

\item {\bf Open access to data and code}
    \item[] Question: Does the paper provide open access to the data and code, with sufficient instructions to faithfully reproduce the main experimental results, as described in supplemental material?
    \item[] Answer: \answerYes{} 
    \item[] Justification: We use publicly available datasets in all experiments. To ensure reproducibility, we will include the complete source code in the supplementary materials in an anonymized form.

\item {\bf Experimental setting/details}
    \item[] Question: Does the paper specify all the training and test details (e.g., data splits, hyperparameters, how they were chosen, type of optimizer, etc.) necessary to understand the results?
    \item[] Answer: \answerYes{} 
    \item[] Justification: We provide details on data splits, hyperparameter settings, and baseline configurations in Appendix~\ref{sec:exp_set}.

\item {\bf Experiment statistical significance}
    \item[] Question: Does the paper report error bars suitably and correctly defined or other appropriate information about the statistical significance of the experiments?
    \item[] Answer: \answerNo{}
    \item[] Justification: Due to the high computational cost associated with training large language models (LLMs), we do not report error bars or variance across multiple runs. However, our evaluation spans a wide range of datasets and task types across multiple domains, and the consistent performance of our method across these diverse settings provides strong empirical support for the robustness and generalization of the proposed approach.

\item {\bf Experiments compute resources}
    \item[] Question: For each experiment, does the paper provide sufficient information on the computer resources (type of compute workers, memory, time of execution) needed to reproduce the experiments?
    \item[] Answer: \answerYes{} 
    \item[] Justification: We report the computer resources in Appendix~\ref{sec:exp_set}.
    
\item {\bf Code of ethics}
    \item[] Question: Does the research conducted in the paper conform, in every respect, with the NeurIPS Code of Ethics \url{https://neurips.cc/public/EthicsGuidelines}?
    \item[] Answer: \answerYes{} 
    \item[] Justification: We confirm that our work complies with the NeurIPS Code of Ethics, including ethical standards for data collection, usage, and experimentation.

\item {\bf Broader impacts}
    \item[] Question: Does the paper discuss both potential positive societal impacts and negative societal impacts of the work performed?
    \item[] Answer: \answerNA{} 
    \item[] Justification: The paper does not discuss societal impacts because it focuses on foundational research in graph learning. The method is generic and does not directly involve application scenarios with immediate societal implications.
    
\item {\bf Safeguards}
    \item[] Question: Does the paper describe safeguards that have been put in place for responsible release of data or models that have a high risk for misuse (e.g., pretrained language models, image generators, or scraped datasets)?
    \item[] Answer: \answerNA{} 
    \item[] Justification: We do not release any data or models that pose high risk, and this work does not involve or introduce such risks.

\item {\bf Licenses for existing assets}
    \item[] Question: Are the creators or original owners of assets (e.g., code, data, models), used in the paper, properly credited and are the license and terms of use explicitly mentioned and properly respected?
    \item[] Answer: \answerYes{} 
    \item[] Justification: We properly cite the original papers corresponding to all models and datasets used in our work. Wherever applicable, we also ensure that we comply with the licenses and terms of use of these assets. For publicly available datasets and pretrained models, we use only those released under permissible licenses (e.g., MIT, CC-BY).

\item {\bf New assets}
    \item[] Question: Are new assets introduced in the paper well documented and is the documentation provided alongside the assets?
    \item[] Answer: \answerYes{} 
    \item[] Justification: Our paper does not release new assets.

\item {\bf Crowdsourcing and research with human subjects}
    \item[] Question: For crowdsourcing experiments and research with human subjects, does the paper include the full text of instructions given to participants and screenshots, if applicable, as well as details about compensation (if any)? 
    \item[] Answer: \answerNA{} 
    \item[] Justification: The paper does not involve crowdsourcing nor research with human subjects.

\item {\bf Institutional review board (IRB) approvals or equivalent for research with human subjects}
    \item[] Question: Does the paper describe potential risks incurred by study participants, whether such risks were disclosed to the subjects, and whether Institutional Review Board (IRB) approvals (or an equivalent approval/review based on the requirements of your country or institution) were obtained?
    \item[] Answer: \answerNA{} 
    \item[] Justification: The paper does not involve crowdsourcing nor research with human subjects.

\item {\bf Declaration of LLM usage}
    \item[] Question: Does the paper describe the usage of LLMs if it is an important, original, or non-standard component of the core methods in this research? Note that if the LLM is used only for writing, editing, or formatting purposes and does not impact the core methodology, scientific rigorousness, or originality of the research, declaration is not required.
    \item[] Answer: \answerYes{}
    \item[] Justification: This paper employs a large language model (Vicuna-7B) as a core component of the proposed encoder-decoder framework. Specifically, the LLM is used as the decoder to generate task-specific outputs conditioned on alignment tokens. The LLM’s generalization ability is integral to the design of our method and plays a central role in achieving zero-shot performance across diverse graph tasks.

\end{enumerate}

\end{document}